%% file: main.tex
\documentclass[11pt,a4paper]{article}

\usepackage{liatuo}          % Li Auto house style
\usepackage{tocloft}
\usepackage{multicol}
\usepackage{adjustbox}
\usepackage{makecell}

\title{ME-Brain-1.0: Memory, Cognition and Action for Evolving Embodied Intelligence}
\author{\textbf{Foundation Model, Li Auto Inc.}}
\shorttitle{MachEmbodied-Brain Technical Report}
\reportdate{September 2026}

\begin{document}

%% ── Title block ─────────────────────────────────────────────────────────────
\makeLiTitle

%% ── Abstract ────────────────────────────────────────────────────────────────
\begin{liabstract}
Current embodied systems largely rely on pretrained capabilities that remain fixed after deployment, limiting their ability to learn from physical interaction. We introduce \textbf{MachEmbodied-Brain (ME-Brain)}, a self-evolving embodied system organized around a closed loop of action execution, experience acquisition, experience evolution, and improved execution. \textbf{Evolvable Memory} consolidates multimodal trajectories into hierarchical, reusable experience; \textbf{Cognitive Core} transforms physical experience into transferable skills; and the \textbf{Action Model} combines event-driven keyframes, EventCell local-world prediction, and action-conditioned memory modulation to focus computation on decision-critical moments, regions, and historical evidence. Together, these modules shift embodied intelligence from \textit{train-and-freeze} to \textit{deploy-and-evolve} without model retraining.

Cognitive Core outperforms the strongest comparison models by \textbf{8.2} and \textbf{9.6 points} on embodied and agent benchmarks. The Action Model achieves \textbf{47.88\%} mean success on RoboMME, a \textbf{3.26-point} improvement over the strongest baseline. On RoboDojo, it reaches a \textbf{21.51} mean Score and \textbf{16.03\%} success rate, exceeding $\pi_{0.5}$ by \textbf{10.10} and \textbf{9.12 points}. On the six-task ME-RealBench, ME-Brain achieves a \textbf{69.5} mean Score and \textbf{66.7\%} success rate, outperforming DM0.5 by \textbf{12.8} and \textbf{11.7 points}, respectively.
\vspace{1em}

\noindent\textit{\textbf{Project Page:}\enspace\url{https://machembodied.com/ME-Brain/ME-Brain-1.0.html}}

\noindent\textit{\textbf{Code Repository:}\enspace\url{https://github.com/MachEmbodied/ME-Brain-1.0}}

\end{liabstract}

\clearpage
%% ── Table of contents ───────────────────────────────────────────────────────
\tableofcontents

\clearpage

%% ── Main sections ───────────────────────────────────────────────────────────
\input{sec/01_introduction}

\input{sec/02_related_work}
\input{sec/03_MachEmbodied-Brain_Framework}
\input{sec/04_Evolvable_Memory}
\input{sec/05_Cognition_Core}
\input{sec/06_Action_Model}
\input{sec/07_Experiments}
\input{sec/08_Qualitative_Examples}
\input{sec/09_Conclusion_and_Future_Works}
%% ── Bibliography ────────────────────────────────────────────────────────────
\bibliography{references}

\clearpage
\input{sec/Contributiors}

%% ── Appendix ────────────────────────────────────────────────────────────────
\appendix

\end{document}

%% file: sec/01_introduction.tex
%% sec/01_introduction.tex

\section{Introduction}
Recent advances in large vision-language models (VLMs) and vision-language-action (VLA) models have substantially improved the perception and control capabilities of embodied systems~\citep{beyer2024paligemma,kim2024openvla,black2024pi0,physicalintelligence2025pi05}. Embodied brains built on agentic architectures have further enhanced manipulation performance in controlled environments~\citep{tan2025roboos,liu2026phyagentos,acebrainteam2026acebrain05}. However, existing embodied brains largely assemble the capabilities of pretrained agentic foundation models, with their perception, planning, and action control capabilities remaining largely fixed once model parameters are frozen after training. When confronted with novel environments, objects, or tasks outside the training distribution, these systems must rely on their previously acquired generalization capabilities, rather than learning and improving through continual interaction with the physical world. This fixed-capability paradigm has two fundamental limitations:

\begin{itemize}
    \item First, a single large-scale training process cannot anticipate all deployment scenarios, making distribution shifts between training data and real-world operating environments difficult to eliminate.
    \item Second, embodied systems should be capable of continual learning and evolution. Robots accumulate substantial operational experience during long-term deployment, including successful executions, failure diagnoses, and environment-specific adaptation strategies. Yet this experience cannot be incorporated back into the models themselves, leaving valuable learning signals underutilized.
\end{itemize}

Overcoming these limitations requires systems to continually self-evolve after deployment, an objective that recent embodied-agent systems have begun to explore through experience reuse, skill reflection, and modular adaptation~\citep{liu2026phyagentos,ju2026embodiskill,chen2026host}. This raises three interdependent questions: how experience is acquired, how experience evolves, and how experience guides action execution. We introduce ME-Brain, an embodied system comprising three core modules—Evolvable Memory, Cognitive Core, and Action Model—that address these questions, respectively. Together, they establish a continual self-evolution loop of action execution → experience acquisition → experience evolution → action execution, enabling robots to autonomously improve through physical interaction without model retraining.

As the experience hub of this loop, Evolvable Memory addresses the fundamental question of where experience comes from. During each task execution, it structurally encodes multimodal streams, including visual observations, actions, and task states, into task-node-level memory summaries grounded in the execution process. It then constructs an external memory graph, with each complete task represented as a parent node and its subtasks and key events represented as linked child nodes. Maintained independently of the model’s context window, this graph overcomes the constraints of static, length-limited memory in conventional approaches~\citep{chhikara2025mem0,li2025memos,shi2025memoryvla}. Building on this representation, the system draws inspiration from human memory to establish a three-tier hierarchy of short-, mid-, and long-term memory, enabling continual experience accumulation through a recording–consolidation–abstraction process. Recent work likewise demonstrates that memory at multiple temporal scales and abstraction levels is essential for long-horizon robot control~\citep{sridhar2025memer,torne2026mem}. Short-term memory records temporally dense multimodal state information during task execution, providing immediate historical context for local action decisions. As the task progresses, the system identifies and extracts state transitions, action outcomes, and anomalous feedback that are critical to task progress, organizing them into salient execution records in mid-term memory to provide task-level context for Cognitive Core. Through cross-task event comparison and outcome analysis, the system further distills consistent success conditions, recurring failure modes, and their applicable contexts from mid-term memory into abstract long-term knowledge, supporting retrieval, transfer, and reuse in subsequent tasks.

Cognitive Core serves as the cognitive engine of experience evolution, addressing how cognition continually drives the refinement of accumulated experience. The central challenge is to transform execution processes in the physical world into reusable behavioral capabilities. This requires both accurate representation and understanding of real-world execution states and action–outcome causality, as well as the ability to consolidate and organize experience into skills and strategies, a direction increasingly explored through hierarchical orchestration, execution feedback, and skill-aware reflection~\citep{tan2025roboos,liu2026phyagentos,li2026roboclaw,ju2026embodiskill}. To this end, Cognitive Core integrates embodied cognition with multimodal agent capabilities. Embodied cognition ensures the quality and interpretability of acquired experience through four core capabilities: embodied scene understanding jointly models physical entities and spatial geometry, enabling memory records to represent real-world physical states; long-horizon planning decomposes complex tasks into executable action sequences, allowing experience to be organized around subtask structures; execution understanding provides a causal basis for experience consolidation by interpreting action-induced state transitions; and diagnosis and replanning support execution recovery while continually writing both successful and failed experiences into Evolvable Memory. Multimodal agent capabilities, in turn, provide the mechanisms for transforming experience into reusable competencies. Multimodal understanding determines the information richness and semantic fidelity of experience representations, influencing subsequent retrieval and cross-task analogy. Agent skill invocation encapsulates executable behaviors as Embodied Skills, while a skill-update protocol consolidates effective strategies into new skills, directly enabling experience-to-skill evolution. Long-horizon task understanding maintains state continuity across execution steps and tasks, retrieves relevant experience, and supports dynamic replanning, making it essential for task-level capability transfer. Finally, reasoning and attribution identify the root causes of execution outcomes and distill actionable experience, determining the quality and potential of experience evolution.

Action Model generates executable robot actions from language instructions and serves as the low-level control component through which behavioral improvements are realized. Contemporary generalist policies formulate this mapping with autoregressive or flow-based VLA architectures~\citep{kim2024openvla,black2024pi0,physicalintelligence2025pi05}, while recent world--action models introduce future visual dynamics as an additional source of action supervision~\citep{ye2026dreamzero,ye2026gigaworldpolicy}. Motivated by the observation that decision-relevant information in robotic manipulation is concentrated at a small number of critical moments, interaction regions, and historical states, we propose Focus-VLWA, an interaction-event-centric model that augments VLA with a world-model expert to jointly model action generation and event-relevant future prediction. The model organizes action keyframes around salient joint movements and gripper-state changes, reducing the learning overhead associated with stationary and repetitive states. Through EventCell, it focuses on future changes within the interaction region and jointly predicts the target pose of the next grasping or placing event, providing local dynamics supervision and event-level goal guidance for action generation. Through action-conditioned memory modulation, it selectively retrieves historical evidence and directly modulates the action expert, avoiding continual expansion of the vision-language context, in contrast to approaches that retain history through explicit memory tokens or multi-scale context~\citep{shi2025memoryvla,torne2026mem,yang2026eventvla}. Together, these designs unify critical moments, interaction regions, and historical evidence within an action-oriented selective modeling framework, concentrating computation on information that directly affects manipulation decisions.

The three modules form a complete, self-driven loop of action execution → experience acquisition → experience evolution → action execution. Each task execution by Action Model produces multimodal trajectories, which Evolvable Memory consolidates into structured experience across its short-, mid-, and long-term hierarchy, transforming both successful execution paths and failure diagnoses into high-quality experience. Grounded in this experience, Cognitive Core uses reasoning and attribution to identify failure causes, multimodal understanding to interpret execution semantics, and long-context management to retrieve historical patterns across tasks. Through its skill invocation mechanism, it drives experience evolution by consolidating repeatedly validated solutions into new skills and translating long-term experiential knowledge into improved planning strategies. The evolved skill library and memory knowledge base then guide Action Model, enabling improved action trajectories in subsequent tasks and initiating another cycle of experience accumulation and capability refinement. In this way, action execution itself becomes a continual source of learning for ME-Brain.

Our contributions are summarized as follows:

\begin{itemize}
    \item We introduce ME-Brain, a continually self-evolving embodied system that establishes a complete, self-driven loop of action execution → experience acquisition → experience evolution → action execution, enabling robots to autonomously improve through physical interaction without model retraining.
    \item We propose Evolvable Memory, which constructs an external memory graph independent of the model’s context window and organizes experience into a human-memory-inspired hierarchy of short-, mid-, and long-term storage. Through a recording–consolidation–abstraction process, it enables continual accumulation and cross-task reuse of multimodal experience, overcoming the constraints of static, length-limited memory in conventional approaches.
    \item We propose Cognitive Core, which unifies embodied cognition and multimodal cognition. The former grounds experience in accurate and interpretable representations of the physical world through embodied scene understanding, long-horizon planning, execution understanding, and diagnosis and replanning. The latter supports continual experience-to-skill evolution through multimodal understanding, agent skill invocation, long-horizon task understanding, and reasoning and attribution.
    \item We propose an interaction-event-centric Action Model that augments VLA with a world-model expert to jointly model actions, local future changes, and event target poses. Through keyframe representations, EventCell-based local prediction, and action-conditioned memory modulation, it focuses on critical moments, interaction regions, and historical evidence, reducing computational redundancy while providing historical support and event-level goal guidance for action generation.
\end{itemize}

\label{sec:intro}

%% file: sec/02_related_work.tex
%% sec/02_related_work.tex

\section{Related Work}
\label{sec:related}

\subsection{Multimodal Agents}
\label{sec:related_multimodal_agent}

Recent multimodal agent models increasingly unify visual-language alignment,
long-context management, reasoning, and tool use \cite{qwen38, glm5, team2026mach, team2026kimi}. The Qwen family has
progressively strengthened these capabilities from Qwen2.5-VL \cite{qwen2.5} to Qwen3.8 \cite{qwen38}.
GLM-5 \cite{glm5} combines agentic interaction, reasoning, and coding within a single
mixture-of-experts architecture and uses asynchronous reinforcement-learning
infrastructure and agent RL to improve long-horizon interaction, tool use, and
self-correction. Kimi K3 \cite{team2026kimi} scales this paradigm with a 2.8-trillion-parameter MoE
architecture and a million-token context window, applying reinforcement learning
across long-horizon coding, general agents, and multimodal reasoning. These
general-purpose agents are primarily trained for interaction in digital spaces
such as text, code, and web pages. They do not systematically model the spatial
structure of the physical world, robot action semantics, or execution-state
monitoring. This limitation motivates the explicit integration of multimodal
agent capabilities with embodied cognition in ME-Brain.

\subsection{Embodied Vision-Language Models}
\label{sec:related_embodied_vlm}

As VLMs have advanced, recent work has transferred general multimodal
understanding to physical AI. RynnBrain \cite{dang2026rynnbrain} supports egocentric understanding,
spatiotemporal localization, physically grounded reasoning, and
perception-aware planning in a unified embodied foundation model. RynnBrain 1.1 \cite{rynnbrain1_1} 
adds contact-point prediction, native 3D spatial perception, and a unified
cross-embodiment action space. Embodied-R1.5 \cite{yuan2026embodied} integrates spatial cognition, task
planning and correction, and embodied pointing through a
Planner--Grounder--Corrector loop. VeBrain-1.5 ~\citep{yang2026vebrain15} reformulates robot control as
MLLM-compatible 2D visual-space tasks and unifies perception, spatial reasoning,
and control through a shared decision interface.
Hy-Embodied-VLM-1.0 \cite{hy_embodied_vlm1_0} organizes physical reasoning around action-related state
understanding, action-transition reasoning, and sequence-adaptive reasoning.
ACE-Brain-0.5 further couples spatial perception, decision making, embodied
interaction, self-monitoring, and self-evolution in a single model
~\citep{acebrainteam2026acebrain05}. Generalist embodied reasoning models also explore multimodal agent capability
and cross-domain transfer. Vesta \cite{bjorck2026vesta} unifies localization, navigation, embodied
question answering, and long-horizon planning with multimodal memory, while
MiMo-Embodied \cite{hao2025mimoembodiedxembodiedfoundationmodel} jointly trains on autonomous driving and indoor embodied AI. Yet
existing systems still tend to emphasize either physical grounding and action
generation or open-ended agent reasoning and skill use. ME-Brain treats the two
capability families as mutually reinforcing objectives and trains them within a
unified Cognitive Core.

\subsection{Embodied Operating Systems}
\label{sec:related_embodied_os}

\textbf{Embodied capabilities and computational infrastructure.}
Embodied AI for scientific discovery relies on both model capabilities and supporting
infrastructure. Prior work highlights edge computing, modular hardware, localized
data pipelines, and open standards for practical deployment, particularly in
resource-constrained
environments~\citep{liu2026embodiedscience,liu2026infrastructurefirst}. Unified
embodied models and world--action modeling connect perception, decision-making, and
execution through shared multimodal interfaces, integrated spatial reasoning and
control, and representations of action-relevant physical states and
outcomes~\citep{yang2026vebrain15,acebrainteam2026acebrain05,kairosteam2026kairos}. At
the computational level, unified inference infrastructure reduces implementation
fragmentation across evaluation, cloud rollouts, edge serving, and on-robot
deployment. Model adapters and shared execution mechanisms preserve model-specific
logic while supporting low-latency interaction and scalable rollout
generation~\citep{wang2026phyai}. These efforts establish model- and deployment-level
foundations, whereas system-level research further addresses heterogeneous capability
orchestration, interaction state maintenance, and execution monitoring.

\textbf{Embodied system orchestration and continual adaptation.}
Hierarchical embodied frameworks connect high-level planning, skill invocation, and
physical execution, exploring cross-embodiment collaboration, shared state, execution
protocols, and outcome verification for multi-stage task coordination and
supervision~\citep{liu2026phyagentos,tan2025roboos}. Complementary embodied agent
harnesses structure the perception--decision--execution loop through action and
observation abstractions, tool interfaces, state maintenance, and structured feedback,
enabling state tracking, feedback-driven replanning, and failure recovery in
long-horizon tasks~\citep{liu2026guava,wang2026harness}. For capability adaptation,
reflection-based methods distinguish deficiencies in skill knowledge from execution
errors to selectively refine procedural knowledge, while demonstration-based methods
align human demonstrations with robot trajectories to rapidly acquire manipulation
behaviors, both without task-specific parameter
updates~\citep{ju2026embodiskill,chen2026host}. Together, these efforts address task
orchestration, closed-loop interaction, and skill refinement, providing complementary
mechanisms for coordinating capability invocation, execution verification, and
experience reuse within an Embodied OS.

%% file: sec/03_MachEmbodied-Brain_Framework.tex
%% sec/03_MachEmbodied-Brain Framework.tex

\section{MachEmbodied-Brain Framework}
\label{sec:framework}

Post-deployment self-evolution in embodied systems requires addressing three
interdependent questions: how experience is acquired, how experience evolves,
and how experience drives action execution. ME-Brain assigns these functions to
three core modules---Evolvable Memory, Cognitive Core, and Action Model---and
organizes them into a self-driven loop of action execution, experience
acquisition, experience evolution, and improved action execution. This design
allows the robot to continue learning from physical interaction without model
retraining.

\begin{figure}[t]
    \centering
    \includegraphics[width=\linewidth]{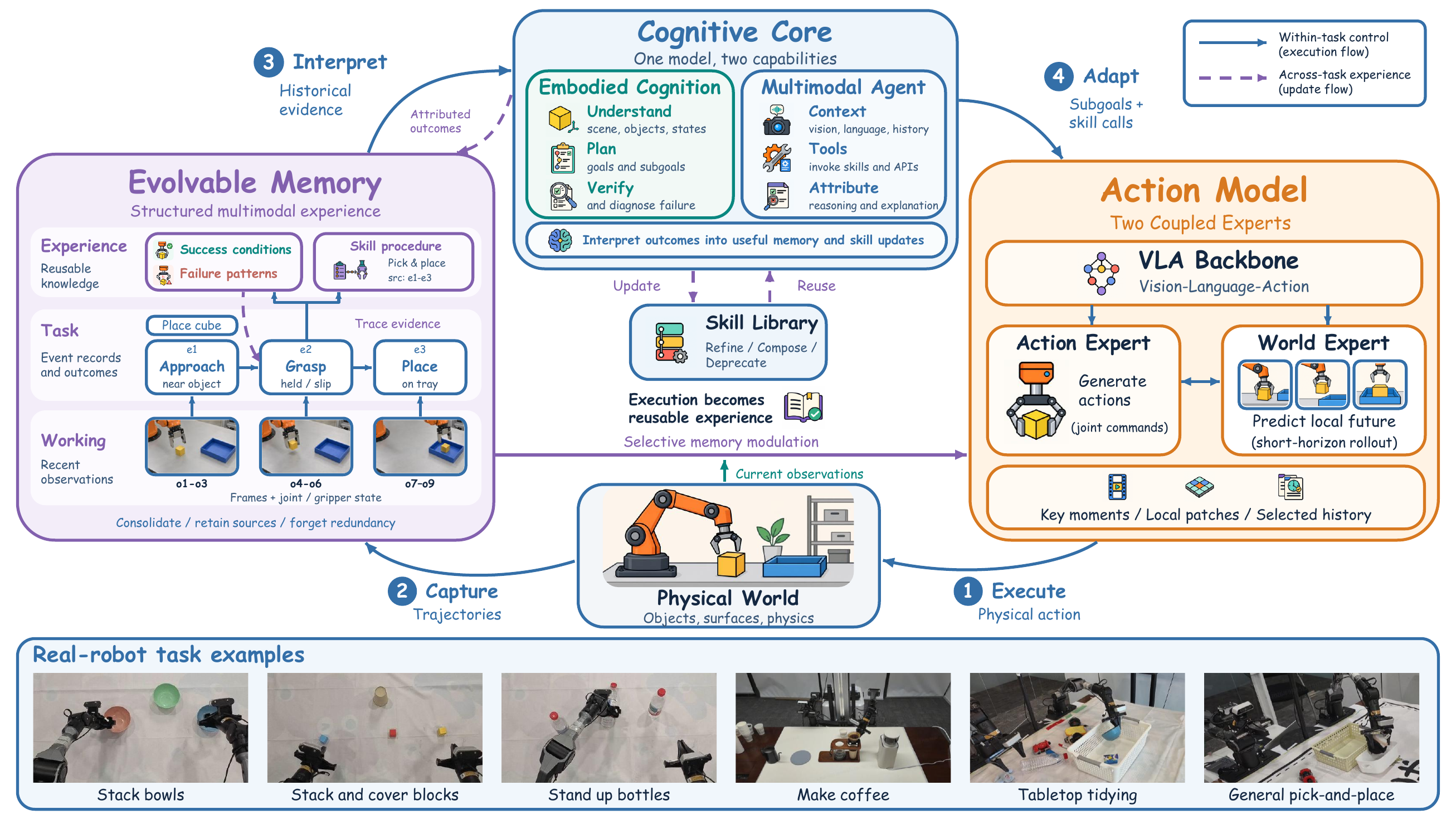}
    \caption{\textbf{Experience-driven self-evolution framework of ME-Brain.}
    Cognitive Core integrates embodied cognition with multimodal agent
    capabilities to plan subgoals, invoke skills, and interpret execution
    outcomes using current observations and historical evidence. Evolvable
    Memory consolidates observations and states into task events and reusable
    experience. Action Model executes the resulting subgoals through the
    coordination of the Action Expert and WM Expert. Together, the three modules
    connect within-task execution with cross-task experience evolution.}
    \label{fig:me_brain_main}
\end{figure}

ME-Brain is designed to move beyond the train-and-freeze paradigm. After initial
training, each task execution becomes an incremental source of system capability.
The framework divides the system into memory, cognition, and action functions:

\begin{itemize}
    \item \textbf{Evolvable Memory} structurally encodes and hierarchically
    consolidates multimodal execution streams into an external memory that is
    independent of the model context window, addressing how experience is
    acquired.
    \item \textbf{Action Model} translates language instructions into executable
    robot actions and serves as the low-level control component through which
    accumulated experience influences execution.
    \item \textbf{Cognitive Core} acts as the system hub, integrating embodied
    cognition with multimodal agent capabilities to interpret, consolidate,
    attribute, and transform experience into reusable skills.
\end{itemize}

The modules form a causal chain within the self-evolution loop. Action Model is
the source of new experience: it executes subgoals issued by Cognitive Core,
translates language intent into physical action sequences, and continuously
produces multimodal trajectories containing perceptual states, action decisions,
and execution outcomes. Evolvable Memory is the consolidation layer: it encodes
and archives these trajectories in an external memory graph, converting a
one-time execution into persistent knowledge that can be retrieved and
transferred. Cognitive Core is the experience refinement layer: it retrieves
historical evidence, attributes success conditions and failure modes, and
consolidates repeatedly validated strategies into new skills that can be invoked
by Action Model.

The output of each module becomes the input to the next, forming a complete
causal chain in which execution produces experience, experience is recorded,
records are transformed into capability, and evolved capability improves future
execution. The autonomous operation of this chain allows ME-Brain to expand its
capability boundary through continued deployment rather than external retraining.

%% file: sec/04_Evolvable_Memory.tex
\section{Evolvable Memory}

Reactive action models such as OpenVLA generate the next action from the current image observation and language instruction, without explicitly conditioning on observation history~\cite{kim2024openvla}. In long-horizon tasks, however, the evolving states of objects, accumulated outcomes of past actions, causes of failure, and dynamic environmental changes often exhibit strong causal dependencies. Under partial observability, different underlying states can produce indistinguishable current observations, leading to state aliasing and limiting policies that treat the current observation as a complete state~\cite{kaelbling1998planning}. Furthermore, embodied systems operating over extended periods must maintain environmental states and task experiences across time from continuous observations, requiring long-term multimodal memory to guide action generation. Maintaining such memory introduces information redundancy and computational overhead, which constrain the temporal coverage, information density, and retrieval efficiency of existing memory approaches. Existing memory and retrieval systems use textual or structured semantic abstractions: SimpleMem consolidates interactions into compact semantic memory units~\cite{liu2026simplemem}, MemoryBank maintains conversation records and hierarchical event summaries~\cite{zhong2024memorybank}, and GraphRAG organizes document information into entity graphs and community summaries~\cite{edge2024local}. MIRIX extends memory to multimodal inputs through specialized memory components~\cite{wang2025mirix}. These designs motivate a complementary requirement for embodied action generation: retaining fine-grained visual, spatial, and action-state information alongside semantic abstractions.

\begin{figure*}[t]
    \centering
    \includegraphics[width=\textwidth]{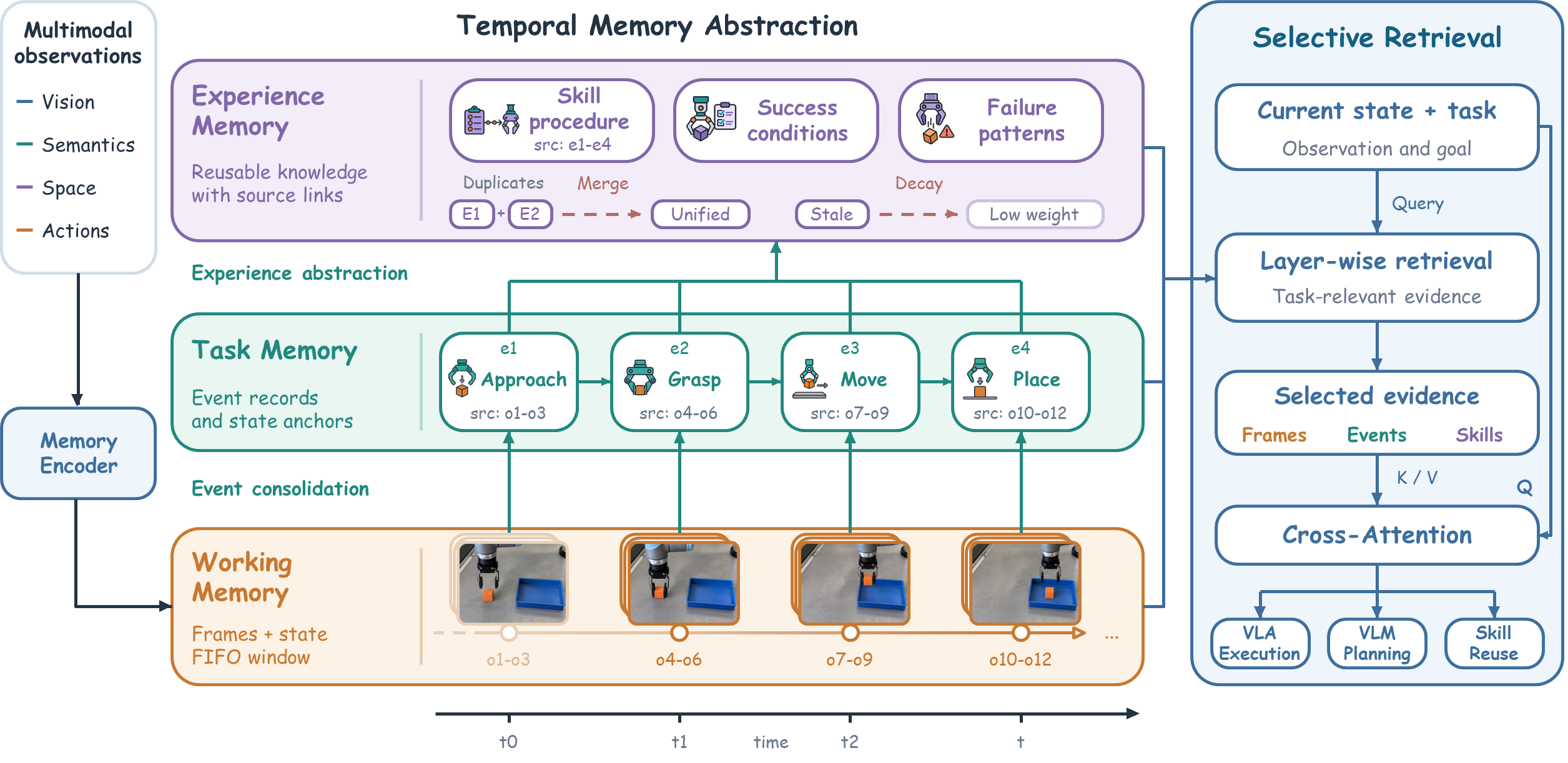}
    \caption{\textbf{Multimodal hierarchical memory and selective access.}
    Multimodal observations are encoded by a memory encoder and written to working memory, consolidated into task memory through event aggregation, and further abstracted into skill procedures, success conditions, and failure patterns. Provenance links preserve associations with supporting evidence across memory levels, while first-in, first-out (FIFO) eviction, duplicate merging, and downweighting of stale information support memory maintenance. The current state and task guide hierarchical retrieval, and the selected evidence is incorporated through cross-attention to support action execution, task planning, and experience reuse.}
    \label{fig:evolvable_memory}
\end{figure*}

To address these limitations, we propose \textit{Evolvable Memory}, an evolvable multimodal memory module for embodied intelligence that combines global causal modeling with local Markovian action modeling. By incorporating key historical information, including spatial locations, inter-object relationships, and execution states at the time of an event, Evolvable Memory provides historical conditions that are difficult to recover from the current observation alone, thereby improving the understanding and execution of long-horizon tasks. As illustrated in Figure~\ref{fig:evolvable_memory}, the memory system progressively organizes multimodal interaction history into hierarchical representations and selectively retrieves information according to the requirements of downstream modules.

To support fine-grained memory over extended interaction horizons, Evolvable Memory incorporates two key mechanisms:

\begin{itemize}
    \item \textbf{Multimodal Embodied Memory Storage.}
    In addition to high-level semantic memories, the module preserves low-level visual and action memories to provide fine-grained guidance for task planning and execution. Hierarchical storage and evolutionary compression consolidate memories into compact, generalizable knowledge, allowing them to persist beyond a single context window and to be continuously updated throughout task execution, ultimately forming a memory knowledge base with no fixed theoretical upper bound. Meanwhile, more efficient memory retrieval enables more efficient memory injection, reducing its computational complexity from $\mathcal{O}(n^2)$ to $\mathcal{O}(n)$ relative to conventional memory injection approaches.

    \item \textbf{Continual Memory Evolution.}
    To support long-term embodied operation, we design a memory evolution mechanism that enables continual accumulation and dynamic updates. Through hierarchical storage, evolutionary compression, and on-demand retrieval and injection, embodied memories are continuously recorded and evolve from the bottom up, progressively consolidating raw observations into events and skill-related experience. As memory evolves, the Action Model and Cognitive Core develop an increasingly detailed understanding of the environment and their own capability boundaries. The former uses its understanding of these boundaries to reduce trial and error, while the latter draws on historical failure cases to identify root causes and resume execution more efficiently. Continual memory evolution enables the overall system to become increasingly proficient over extended operation.
\end{itemize}

\subsection{Memory Storage and Evolution}

\subsubsection{Multimodal Embodied Memory Storage}

Memory systems for embodied tasks must support both high-level task planning and fine-grained action generation. High-level semantics can summarize task objectives, execution stages, and successful or failed experiences, whereas concrete manipulation additionally depends on object locations, inter-object relationships, state changes before and after actions, and the temporal order in which these changes occur. Retaining only textual summaries of events can therefore result in the loss of execution-relevant information. Accordingly, our memory system preserves visual observations, spatial states, and action information alongside high-level semantic memories, enabling historical experience to guide task understanding and concrete execution at different levels of granularity.

To construct these memory representations, we introduce an \textit{Embodied Model} specifically adapted to memory-related tasks as a specialized memory expert. The model receives multimodal observations and action states collected during task execution, extracts semantic information from key events, updates embodied state representations, and generates structured memory nodes. Each node organizes retrievable textual semantics together with the associated visual context, object locations, inter-object relationships, and action outcomes, preserving the execution-level context underlying summarized experience. Through the links between nodes in the hierarchical memory structure, the system can use high-level semantics for planning while accessing fine-grained historical conditions relevant to the current operation.

Consider the task of picking up a cup, replacing the tablecloth, and then returning the cup to its original position. The final placement action depends on the cup's location at the beginning of the task. A semantic summary such as ``the cup was originally on the table'' captures the general event but does not specify a sufficiently precise placement target. A multimodal memory node preserves both the original spatial location and the associated visual context, allowing the system to retrieve the relevant historical state when executing the instruction to ``return the cup to its original position'' and to provide placement conditions to the base action model through memory injection. This example illustrates the complementary roles of semantics and embodied states: the former supports the interpretation of task requirements, while the latter provides the concrete information needed to generate actions that satisfy those requirements.

\subsubsection{Memory Evolution}

Evolvable Memory supports memory retention over month-scale time horizons, allowing historical information to persist across individual tasks and beyond model context windows. During long-horizon tasks, object states, action outcomes, and environmental changes continuously accumulate, collectively forming the historical conditions required for subsequent decisions. Relying solely on current observations and a limited context buffer makes it difficult to maintain this information over extended periods. Evolvable Memory therefore combines hierarchical storage, evolutionary compression, and on-demand retrieval and injection to retain long-term history while accessing the information required for current decisions, extending the temporal scope of memory while controlling maintenance and inference overhead.

Across temporal scales, Evolvable Memory organizes information into short-term, mid-term, and long-term memory and progressively consolidates information generated during execution through evolutionary compression. Short-term memory retains high-frequency, temporally dense multimodal states, providing recent history for local action execution. Mid-term memory aggregates key events and relationships between execution stages, providing task-level context for planning. Long-term memory stores abstracted success conditions, failure patterns, and semantic experience, supporting retrieval and reuse across tasks. As execution progresses, short-term records are consolidated into mid-term events, and experiences with reuse potential are further abstracted into long-term memory. Mid-term and long-term nodes retain links to the underlying fine-grained information, enabling the system to form compact experience representations while preserving access to specific execution conditions.

In terms of memory organization, Evolvable Memory represents each complete task as a parent node, with subtasks and key events stored as associated child nodes, forming an external memory structure maintained independently of the model's context window. Parent nodes summarize task objectives and overall execution, while child nodes record specific operations, local state changes, and success or failure outcomes. This structure allows the system to trace the relationships between an outcome and the relevant operations and environmental states along the task hierarchy. When historical conditions cannot be obtained directly from current observations, the system retrieves the relevant information through associated nodes, supporting cross-stage planning, action execution, and failure recovery.

To efficiently incorporate continuously accumulated memories into current decisions, Evolvable Memory selects the retrieval level according to the requirements of each downstream module and then locates relevant nodes within that level. The action model retrieves fine-grained states from short-term memory based on spatiotemporal proximity and the objects involved in the current operation. The task planning module retrieves key events and multimodal features from mid-term memory based on task and subtask relationships. The experience summarization module uses semantic similarity to identify reusable experience in long-term memory. This hierarchical retrieval mechanism selects decision-relevant information from the complete history, reducing indiscriminate scanning and the injection of irrelevant content.

\subsection{Edge Deployment, Privacy, and Security}

Long-term embodied memory continuously records environmental observations, user behavior, and task histories, and its deployment must therefore account for both local resource constraints and data privacy requirements. In home, in-vehicle, and industrial settings, on-device memory processing can reduce the transmission of sensitive information between devices and the cloud while shortening the access path for memory generation and retrieval. Evolvable Memory therefore adopts on-device operation as the deployment foundation for long-term memory, allowing its capabilities to directly support local task execution. Built on \textit{Cognitive Core}, Evolvable Memory has been ported to M100 and optimized for this platform, supporting on-device memory generation, storage, evolution, and retrieval. This deployment enables memory-related processing to be performed locally, providing a foundation for maintaining sensitive data on the device. Combined with hierarchical storage and on-demand access, the system organizes and uses memories across different temporal scales on edge devices, supporting long-term embodied applications in real-world environments.
% 03_MachEmbodied-Brain_Framework

%% file: sec/05_Cognition_Core.tex
%% sec/05_Cognition Core.tex

\section{Cognitive Core}
\label{sec:cognitive_core}

Cognitive Core provides the cognitive capabilities of ME-Brain and coordinates
with Evolvable Memory and Action Model to form a complete embodied system \cite{yang2025agentic}. We
develop a balanced training approach that jointly trains embodied-cognition and
multimodal-agent experts through SFT and RL \cite{grpo}, followed by MOPO distillation  to
balance the two capability families. This design allows the model to develop
physical understanding and intelligent decision making within a unified model.

As the central hub of ME-Brain, Cognitive Core retrieves historical evidence
from Evolvable Memory to improve current decisions and writes causally attributed
execution outcomes back into memory to drive continued experience accumulation.
It also sends subgoals and tool or skill invocation instructions to Action Model
and injects recovery strategies when execution deviates from the plan. The three
modules therefore form a closed loop of cognition, memory, and action that
extends self-evolution from individual executions to experience shared across
tasks.

\begin{figure}[t]
    \centering
    \includegraphics[width=\linewidth]{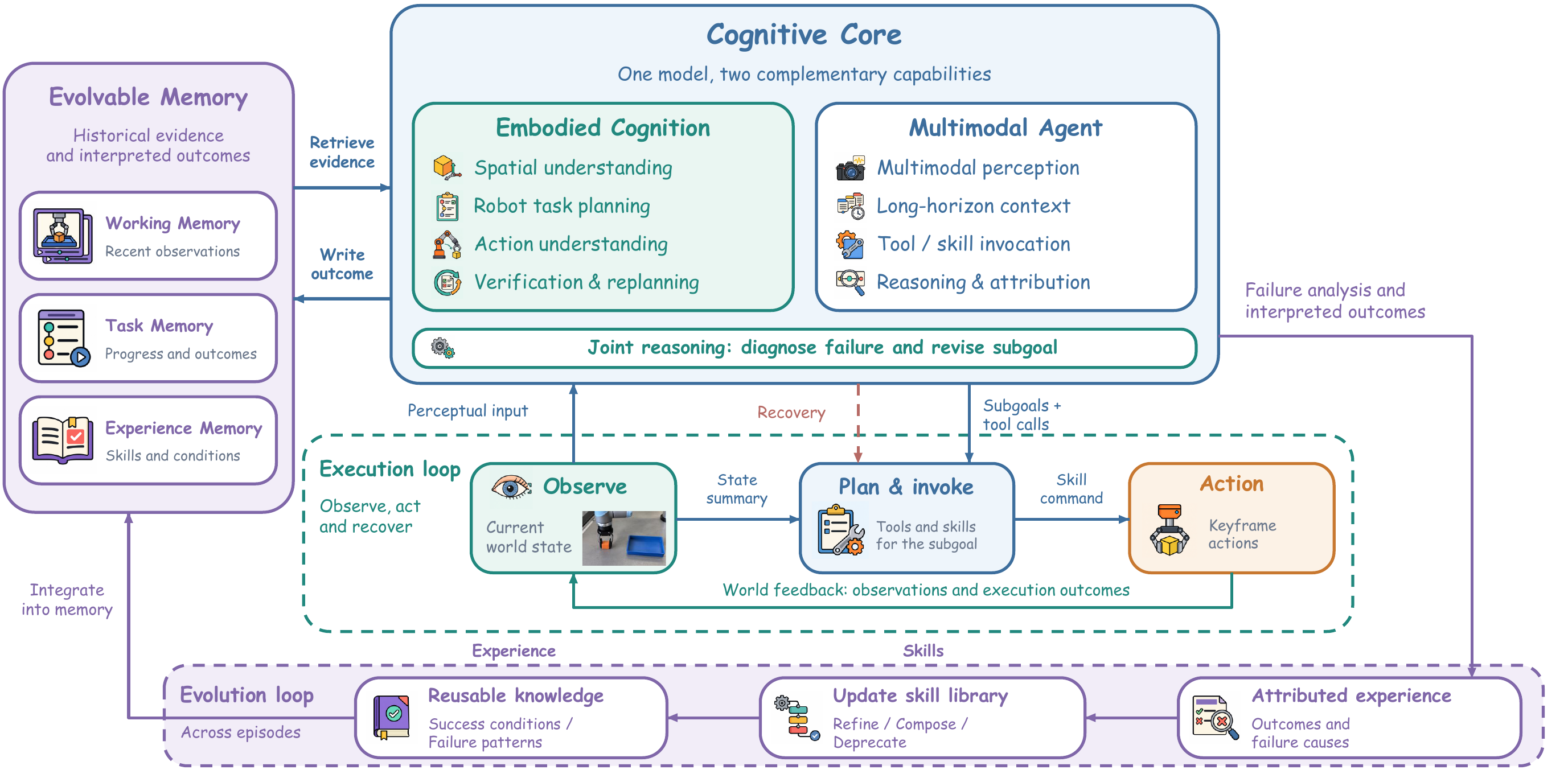}
    \caption{\textbf{Execution and evolution loop of Cognitive Core.}
    The unified cognitive model combines embodied cognition with multimodal
    agent capabilities to interpret the current state using historical evidence,
    plan subgoals, and invoke skills, while Action model executes the resulting
    actions. Observation feedback supports verification, failure diagnosis, and
    replanning. Causally attributed outcomes support skill updates and experience
    abstraction and are written back to memory for future reuse.}
    \label{fig:cognitive_core}
\end{figure}

\subsection{Unified Capabilities}

\subsubsection{Embodied Cognitive Capability}

Embodied cognition is organized around the physical execution loop. Embodied
scene understanding constructs an actionable state representation from
perception; embodied task planning decomposes high-level objectives into
executable subtasks; embodied action understanding relates actions to changes in
the physical state; and verification and replanning diagnose failures and revise
the plan. Together, these capabilities form an observe-plan-act-replan loop
for continuous and robust long-horizon execution.

\begin{enumerate}
    \item \textbf{Embodied scene understanding.}
    A conventional VLM can describe visible content, whereas Cognitive Core must
    additionally determine what is actionable, where to grasp, and where an
    object should be placed. It connects visual semantics, spatial geometry, and
    robot affordances to form an action-oriented state representation. Given an
    observation $o_t$ and instruction $l$, the model produces an
    affordance-aware task decomposition
    \begin{equation}
        \mathcal{A}(o_t,l)
        = \{(r_i,\mathrm{aff}_i,\mathrm{pos}_i)\}_{i=1}^{N},
    \end{equation}
    where $r_i$ describes an actionable region, $\mathrm{aff}_i$ denotes its
    affordance type, and $\mathrm{pos}_i$ is a 2D image coordinate.

    \item \textbf{Embodied task planning.}
    The planner decomposes a high-level instruction into executable and
    verifiable subtasks while jointly considering the objective, environment
    state, available skills, and execution feedback. Given instruction $l$,
    observation $o_t$, robot state $s_t$, and memory $M$, it generates or updates
    a sequence of subtasks:
    \begin{equation}
        \{g_1,g_2,\ldots,g_n\}
        = f_{\mathrm{plan}}(l,o_t,s_t,M).
    \end{equation}
    Each $g_i$ specifies an operation intent, target object, execution
    constraints, and observable completion conditions for Action Model.

    \item \textbf{Embodied action understanding.}
    Beyond predicting actions, the model captures the causal relation between an
    action and its effect on the environment. Given state $s_t$ and proposed
    action $a_t$, it predicts a semantic description of the successor state,
    \begin{equation}
        p(s_{t+1}\mid s_t,a_t),
    \end{equation}
    including object displacement, contact changes, and task-stage progression.
    For example, completing a grasp requires observing gripper closure, object
    lift, and the absence of slippage, rather than merely issuing a grasp command.
    Cognitive Core uses this interpretation to select the next subtask or trigger
    recovery.

    \item \textbf{Verification and replanning.}
    When the current observation $o_t$ violates an expected postcondition, the
    model first identifies the failure cause and then updates the task-progress
    graph and produces a corrected plan. The diagnosis is also written to memory
    as context for future reasoning. Diagnosis quality directly determines the
    quality of the failure experience stored in Evolvable Memory.
\end{enumerate}

\subsubsection{Multimodal Agentic Capability}
Multimodal agent capabilities turn experience into reusable behavior. Multimodal
understanding determines the semantic quality of each experience record;
long-horizon task understanding maintains coherent state across steps and tasks;
tool and skill invocation consolidates experience into reusable composite skills;
and reasoning and attribution provide the basis for extracting useful experience
from execution outcomes. Together, they form a progression from experience
acquisition to capability evolution.

\begin{enumerate}
    \item \textbf{Multimodal understanding.}
    The model extracts manipulation semantics from observations, interprets task
    intent from language, and aligns visual evidence with language reasoning.
    This capability determines the information density of memory nodes and the
    reliability of subsequent retrieval and cross-task analogy.

    \item \textbf{Long-horizon task understanding.}
    The system maintains coherent task state across multiple subgoals and across
    tasks instead of making isolated single-step decisions. A longer retrievable
    history supports broader experience consolidation and the transfer from
    solving one task instance to solving a class of related tasks.

    \item \textbf{Tool and skill invocation.}
    Atomic embodied capabilities are exposed through semantic interfaces that
    can be invoked on demand. Rather than directly producing low-level action
    sequences, the agentic loop orchestrates embodied tools and skills, while an
    Embodied Skill Library stores validated composite skills for cross-task reuse.
    A continual evolution protocol refines preconditions, composes repeatedly
    successful atomic sequences into new skills, and replaces skills that fail
    persistently.

    \item \textbf{Reasoning and attribution.}
    After an execution failure, the system determines whether the cause lies in
    perception, planning, or action precision and generates a targeted correction.
    Stronger attribution improves both the labels attached to failure examples in
    Evolvable Memory and the decisions made by the skill-update protocol, thereby
    setting the upper bound on experience evolution.
\end{enumerate}

\subsection{Collaboration with Other Modules}

Cognitive Core and Evolvable Memory form a bidirectional read--write loop. During
reading, Cognitive Core retrieves key nodes from mid-term event memory using
subtask relationships and matches reusable success conditions and failure
patterns from long-term experience memory using semantic similarity. During
writing, it interprets each execution outcome and produces causally annotated
records describing the failure cause, state-transition process, and subtask
completion status, which are written to the appropriate memory level. Cognitive
Core is therefore both a consumer of memory and a primary producer of
high-quality memory.

Collaboration between Cognitive Core and Action Model operates at two levels.
For execution, the \textit{Observe} node sends the current world state to Cognitive Core;
the embodied-cognition component performs spatial and state analysis, and the
multimodal-agent component converts the result into subgoals and tool or skill
calls for the \textit{Plan} \& \textit{Invoke} node. Action Model then executes the corresponding keyframe action sequence. For closed-loop recovery, action outcomes and new
observations continuously return to the Observe node. Verification and replanning
compare this feedback with expected outcomes, diagnose deviations, and inject a
targeted recovery strategy into \textit{Plan} \& \textit{Invoke} without restarting the complete
task. This loop preserves execution continuity and robustness in unexpected
states.

%% file: sec/06_Action_Model.tex
%% sec/06_Action Model.tex

\section{Action Model}
\label{sec:action_model}

Many vision-language-action (VLA) models generate an action sequence from the
current visual observations $\mathbf O_t$, robot state $\mathbf s_t$, and
language instruction $\ell$, following a policy of the form
$\pi_\theta(\mathbf O_t,\mathbf s_t,\ell)$~\citep{black2024pi0,physicalintelligence2025pi05}.
World-action models extend this formulation by jointly predicting visual futures
and actions~\citep{ye2026dreamzero}, while memory-augmented policies incorporate
temporal context~\citep{shi2025memoryvla,torne2026mem}.
These extensions can introduce substantial redundancy when computation is
allocated across densely sampled action sequences, full future images, or
unfiltered observation histories. Yet task-relevant information in robotic
manipulation is unevenly distributed: many consecutive frames contain little
motion, large image regions are unrelated to the immediate interaction, and only
a small subset of historical states determines the next action. Processing these
inputs without distinguishing their relevance consumes computation and can
dilute the signals associated with contact, grasping, and placement.

\begin{figure}[t]
    \centering
    \includegraphics[width=\linewidth]{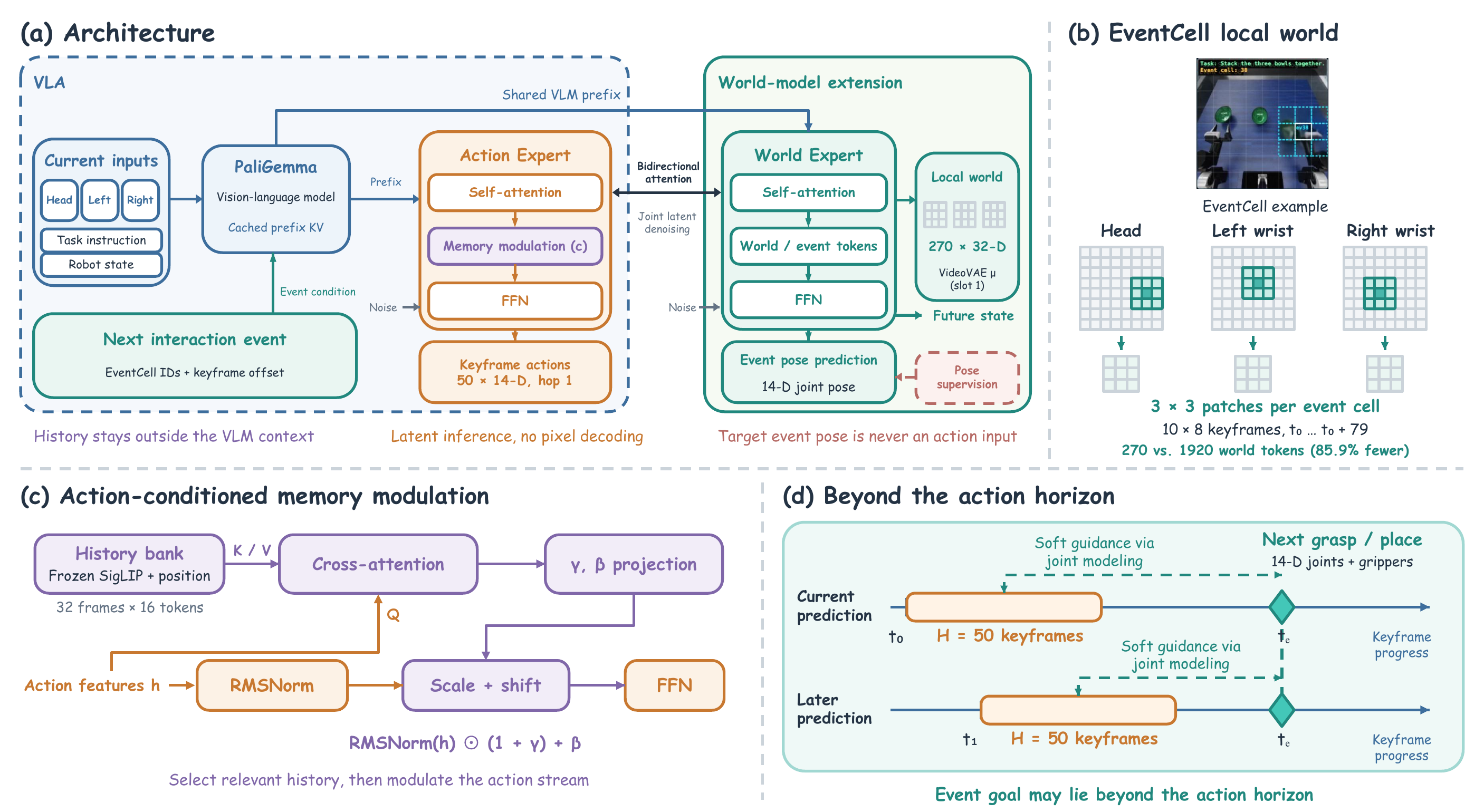}
    \caption{\textbf{Focus-VLWA architecture and key mechanisms.}
    (a) A VLA and a world-model expert share a vision-language prefix and exchange
    information through bidirectional attention for joint latent-space denoising.
    (b) EventCell restricts future-world prediction to multi-view interaction
    neighborhoods.
    (c) Action features query an external history bank and modulate normalized
    action representations without extending the backbone context.
    (d) Fixed-length action windows advance with replanning, while the next event
    target may lie beyond the current horizon and provide guidance through joint
    modeling. Ground-truth event poses are supervision targets, not action inputs.}
    \label{fig:action_model}
\end{figure}

Motivated by this observation, we introduce \textbf{Focus-VLWA}, an
interaction-event-centric \emph{Vision-Language-World-Action} model.
Its central principle is to allocate modeling capacity to the key moments,
interaction regions, and historical evidence associated with the next physical
interaction, rather than treating all temporal and spatial positions alike.
As shown in Figure~\ref{fig:action_model}, Focus-VLWA extends a VLA with a
world-model expert and an action-conditioned memory pathway.

Historical evidence and future-world representations are incorporated without
simply lengthening the model's input and output sequences. Temporally, dense
control trajectories are represented by keyframes corresponding to salient
physical changes. Spatially, future prediction is restricted to neighborhoods
around the next interaction event. For memory, action-conditioned retrieval
selects relevant historical evidence without expanding the vision-language
backbone's context. Together, these mechanisms connect the historical evidence
needed for the current decision, the location and temporal offset of the next
interaction, and the motion required to approach its target state.

\subsection{Event-Centric World-Action Modeling}
\label{sec:event_centric_world_action}

Focus-VLWA adopts the vision-language backbone of
$\pi_{0.5}$~\citep{physicalintelligence2025pi05} to encode the current task state.
The model receives current images from the head, left-wrist, and right-wrist
cameras, the task instruction, and discretized robot state.
PaliGemma~\citep{beyer2024paligemma} produces contextual representations of scene
semantics, manipulation targets, and proprioceptive state. Historical memory
provides visual evidence from the past, while the world-model expert predicts
future changes associated with the next interaction event.

We represent the next interaction event as
\begin{equation}
    e_t = \left(k_t^e, c_t^H, c_t^L, c_t^R,
                 \Delta_t, \mathbf a_t^e\right),
    \label{eq:action_event}
\end{equation}
where $k_t^e$ is the keyframe of the next event, such as a grasp, contact, or
placement; $c_t^H$, $c_t^L$, and $c_t^R$ identify its spatial cells, termed
\emph{EventCells}, in the three camera views; $\Delta_t$ is the number of keyframes
remaining until the event; and $\mathbf a_t^e\in\mathbb R^{14}$ is the target
joint-and-gripper configuration at that event.

The event serves as an alignment anchor for world prediction and action
generation. EventCell specifies where future-world modeling should focus, the
event offset situates the current state on the event timeline, and the event
pose provides the physical target state. World and action predictions are thus
aligned around a shared interaction instead of an undifferentiated visual
future. The event target may lie beyond the fixed action horizon
(Figure~\ref{fig:action_model}(d)), providing goal guidance through joint modeling
without requiring the current action chunk to reach the event immediately.

In the current implementation, EventCell IDs are included in the action model's
prompt as structured conditions. The ground-truth event pose $\mathbf a_t^e$ is
used only for supervision and is not supplied to the action expert. We denote
the event-conditioning information available to the model by $\tilde e_t$, to
distinguish it from the complete event annotation $e_t$ in
Equation~\eqref{eq:action_event}. In particular, $\tilde e_t$ excludes
$\mathbf a_t^e$. The action expert must infer the motion toward the event from
the current observations, task semantics, historical evidence, and available
event conditions, rather than reading the target pose directly. An action event
here is a low-level physical event defined by robot motion and interaction-state
changes, not a high-level semantic event with a human-written description.
Event annotations and conditioning should not themselves be interpreted as an
online event-detection capability.

Along the temporal dimension, Focus-VLWA predicts event-driven keyframe actions
instead of densely sampled control targets at a fixed rate. Adjacent targets in
a dense trajectory often differ only slightly, so a fixed-length prediction
window can be dominated by stationary states or small-amplitude motion and
cover limited task progress. Waypoint-based imitation learning has previously
shown the value of compact trajectory representations for reducing the learning
horizon~\citep{shi2023awe}. Focus-VLWA uses joint and gripper changes to organize
the action sequence on a keyframe timeline shared with future-world supervision.

Specifically, a new action keyframe is retained when a robot joint changes by
more than $1^\circ$ or a gripper displacement exceeds $5\,\mathrm{mm}$.
This resampling compresses the dense control trajectory into a sequence of
salient physical changes. Given the current keyframe index $t_0$, the model
predicts
\begin{equation}
    \hat{\mathbf J}_{t_0} =
    \left[\hat{\mathbf j}_{t_0},\hat{\mathbf j}_{t_0+1},
          \ldots,\hat{\mathbf j}_{t_0+49}\right],
    \qquad \hat{\mathbf j}_i\in\mathbb R^{14}.
    \label{eq:keyframe_actions}
\end{equation}
These 50 positions represent consecutive keyframe targets, not 50 motor commands
at a fixed temporal sampling interval. Each training sample is anchored at a
current keyframe, with the corresponding action supervision
\begin{equation}
    \mathbf J_{t_0}
    = \left[\mathbf j_{t_0},\ldots,\mathbf j_{t_0+49}\right]
    = \mathbf j_{14}[t_0:t_0+50],
    \label{eq:keyframe_supervision}
\end{equation}
where the slice uses an exclusive upper bound. Removing stationary and repeated
states concentrates supervision on meaningful changes in robot configuration
and interaction state. For a fixed output length, the resulting representation
can cover more task progress and a longer elapsed-time span, linking the motion
stages leading toward the next grasp or placement.

Future-world supervision uses the same keyframe timeline. We divide keyframes
$t_0$ through $t_0+79$ into ten world chunks, each containing eight consecutive
keyframes:
\begin{equation}
    [t_0+8i,\ t_0+8i+7], \qquad i=0,\ldots,9.
    \label{eq:world_chunks}
\end{equation}
Instead of making a dense, stride-one future-video prediction at every keyframe,
the world branch uses ten temporally compressed representations to cover 80
keyframes. The action branch predicts consecutive keyframe configurations at
hop one, while the world branch describes environmental changes at a lower
temporal density. Both are aligned to physical task progress rather than a
frame-by-frame replay of the original control stream.

Along the spatial dimension, Focus-VLWA introduces \textbf{EventCell local-world
prediction}. Video-based world-action models such as DreamZero jointly model
future visual observations and actions~\citep{ye2026dreamzero}.
Recent work also explores action-centered modeling to reduce the inference
overhead of video prediction~\citep{ye2026gigaworldpolicy}.
Our design focuses specifically on the spatial support of world supervision:
for manipulation, grasp points, contact regions, gripper states, and target
placement locations are often more informative than background regions far
from the interaction.

We partition the visual latent grid of each camera view into $8\times8$ cells
and locate the EventCell associated with the next interaction. Rather than
predicting the full grid, the world-model expert predicts only the
$3\times3$ neighborhood centered on that cell. Each future world chunk therefore
contains 27 local visual tokens across the head and two wrist views, with nine
tokens per view. The head view provides scene context around the event, while
the wrist views capture local details near the grippers and contact regions.
Each patch is represented by a 32-dimensional vector from temporal slot 1 of the
VideoVAE posterior mean $\mu$. Across ten world chunks, the local-world target
contains 270 such tokens.

Under the same single-temporal-slot representation, predicting the complete
$8\times8$ grid for all three views and ten chunks would require
\begin{equation}
    N_{\mathrm{full}}=10\times3\times64=1920,
    \qquad
    N_{\mathrm{local}}=10\times3\times9=270.
    \label{eq:world_token_budget}
\end{equation}
EventCell thus reduces the number of world tokens by approximately $85.9\%$.
This is a reduction in representation length, not a measured end-to-end latency
reduction. Unlike uniform downsampling of the full future scene, the spatial
selection is tied to the next interaction event, preserving localized changes
relevant to the action while limiting the world-modeling budget.

The action and world-model experts jointly model keyframe actions, local-world
tokens, future proprioceptive states, and event targets. The current camera
views, task instruction, and robot state form a shared vision-language prefix;
the experts exchange information through bidirectional attention during latent
denoising. With historical memory $\mathbf M_{t_0}$ and event conditions
$\tilde e_{t_0}$, this joint predictor is written as
\begin{equation}
    F_\theta\!\left(
        \mathbf O_{t_0},\mathbf s_{t_0},\ell,
        \mathbf M_{t_0},\tilde e_{t_0}
    \right).
    \label{eq:joint_world_action}
\end{equation}
Its training objective combines
\begin{equation}
    \mathcal L =
    \lambda_a\mathcal L_{\mathrm{action}}
    +\lambda_w\mathcal L_{\mathrm{local\text{-}world}}
    +\lambda_s\mathcal L_{\mathrm{state}}
    +\lambda_e\mathcal L_{\mathrm{event}},
    \label{eq:focus_vlwa_loss}
\end{equation}
where $\mathcal L_{\mathrm{action}}$ supervises future keyframe actions,
$\mathcal L_{\mathrm{local\text{-}world}}$ supervises the VideoVAE latents around
EventCells, $\mathcal L_{\mathrm{state}}$ supervises future proprioceptive states,
and $\mathcal L_{\mathrm{event}}$ promotes consistency among event locations,
temporal offsets, and target configurations.

This formulation makes local dynamics prediction part of action learning.
In addition to matching demonstration trajectories, action representations are
trained jointly with changes in object motion, contact relationships, and gripper
states near the next interaction. The world branch therefore aims to predict
action-relevant local consequences, rather than reconstruct a visually complete
future video as an independent end task.

\subsection{Action-Conditioned Memory Modulation}
\label{sec:action_memory_modulation}

EventCell focuses prediction on future interaction regions, but partially
observable tasks also require evidence that is no longer visible. A target may
be occluded, an operation may need to be repeated a specified number of times,
or an object may need to be returned to its initial position. Identical current
observations can therefore require different actions depending on the preceding
trajectory. Memory-augmented VLA methods address related temporal dependencies
through explicit memory or multi-scale historical
representations~\citep{shi2025memoryvla,torne2026mem}.

A straightforward way to expose history to a policy is to concatenate historical
visual tokens with the current vision-language context. Under dense attention,
increasing this token history increases the backbone's attention workload and
context storage; repeatedly processing the enlarged context can also increase
inference latency. Focus-VLWA instead maintains an independent visual memory
bank and lets the action expert query it directly, leaving PaliGemma's main
context unchanged.

For each action prediction, we uniformly sample up to 32 head-camera frames from
the beginning of the trajectory to the current time. A frozen
SigLIP encoder~\citep{zhai2023siglip}, followed by spatial pooling, produces 16
visual tokens per frame, giving at most 512 historical tokens per read.
Each visual feature is concatenated with temporal and two-dimensional spatial
position embeddings and projected to a 1024-dimensional memory representation:
\begin{equation}
    \mathbf m_{i,j} = f_{\mathrm{mem}}\!\left(
        \left[\mathbf v_{i,j};\mathbf e_i^{\mathrm{time}};
              \mathbf e_j^{\mathrm{space}}\right]
    \right),
    \label{eq:memory_representation}
\end{equation}
where $i$ indexes historical frames and $j$ indexes pooled spatial positions.
These tokens form $\mathbf M_t$ but are not appended to the current image and
language prefix.

At layer $l$ of the action expert, each future action position provides a query,
while the historical memory provides keys and values:
\begin{equation}
    \mathbf c^l = \operatorname{Attn}\!\left(
        Q_l\mathbf h^l,\ K_l\mathbf M_t,\ V_l\mathbf M_t
    \right).
    \label{eq:action_memory_attention}
\end{equation}
The attention weights select evidence relevant to the evolving action
representation. Different action positions can attend to different historical
states, such as an object's initial location, the most recent contact, or evidence
of previously completed repetitions. The retrieved context is projected to
feature-wise scale and shift parameters,
\begin{equation}
    \left(\boldsymbol\gamma^l,\boldsymbol\beta^l\right)
    = g_l(\mathbf c^l),
    \label{eq:memory_modulation_parameters}
\end{equation}
which modulate the normalized action features after self-attention and before
the feed-forward network:
\begin{equation}
    \tilde{\mathbf h}^l
    = \operatorname{RMSNorm}(\mathbf h^l)
      \odot\left(1+\boldsymbol\gamma^l\right)
      +\boldsymbol\beta^l.
    \label{eq:memory_modulation}
\end{equation}
This operation uses RMSNorm~\citep{zhang2019rmsnorm} and a feature-wise affine
conditioning form related to FiLM~\citep{perez2018film}. Our design derives the
modulation parameters from action-conditioned retrieval over a separate visual
history bank and applies them within the action expert, rather than adding
historical tokens to the backbone context.

Cross-attention determines which evidence to retrieve, while the scale and
shift parameters control how that evidence changes the action features.
Memory thus influences action generation at different network depths and future
action positions, rather than serving only as a static input prompt.

Let $N_c$, $N_m$, and $N_a$ denote the numbers of current-context, memory, and
action tokens. For fixed feature dimensions and network depth, concatenating
memory into a densely attended backbone context yields an attention-score
complexity of
\begin{equation}
    O\!\left((N_c+N_m)^2\right).
    \label{eq:concatenation_complexity}
\end{equation}
Keeping the backbone context unchanged and reading memory through the action
expert instead gives
\begin{equation}
    O(N_c^2)+O(N_aN_m),
    \label{eq:modulation_complexity}
\end{equation}
omitting costs that do not vary with the memory length, such as action
self-attention at a fixed horizon. Thus, for fixed $N_a$, the additional memory
attention scales linearly with the number of retrieved tokens. This comparison
concerns attention computation, not a claim that total system latency follows
the same asymptotic ratio. In the current implementation, the 32-frame sampling
cap bounds the token budget while allowing the sampled history to span a longer
trajectory.

During training, historical memory supplies past evidence and local-world
supervision constrains future consequences. Future-world latents, robot states,
and actions are trained with a joint flow-matching
objective~\citep{lipman2023flowmatching}. Asynchronous noise times expose the
model to modalities at different stages of denoising. The vision-language
backbone and VideoVAE remain frozen; training updates the action and world-model
experts, memory connectors, and newly introduced event and action prediction
modules. This preserves the pretrained vision-language and video priors.

At deployment, the vision-language prefix and historical memory representations
are computed once per observation update and reused across action-denoising
steps. The action-serving path operates in latent space without decoding future
video pixels. EventCell latents still participate in joint world-action
inference, but the system avoids pixel-level reconstruction of the full scene
and prediction of spatial regions outside the selected neighborhoods. Skipping
pixel decoding therefore does not mean removing the world-model expert from
action inference.

In summary, Focus-VLWA organizes action generation around selective computation
for physical interactions. Keyframe actions reduce temporal redundancy,
EventCell concentrates future-world modeling on local interaction regions, and
action-conditioned modulation retrieves relevant history without continually
expanding the vision-language context. Historical evidence, local future
changes, and motion toward the next event are combined in a single
vision-language-world-action framework, with computation directed toward the
information that matters for control.

%% file: sec/07_Experiments.tex
%% sec/07_Experiments.tex
%% Synchronized with v3-zh.md on 2026-09-21.

\section{Experiments}
\label{sec:experiments}

\subsection{Evaluation of Cognitive Core}
\label{sec:cognitive_core_evaluation}
\subsubsection{Embodied Benchmark}

For ease of comparison, we define the model alias of Cognitive Core as ME-VLM. The embodied evaluation spans 26 benchmarks covering physical and spatial
understanding, task planning, action execution, and execution correction. We
compare ME-VLM at 35B-A3B and 4B scales with Hy-Embodied 30B-A3B \cite{hy_embodied_vlm1_0},
RynnBrain1.1 9B \cite{rynnbrain1_1}, Mimo-Embodied 7B \cite{hao2025mimoembodiedxembodiedfoundationmodel}, and PhysBrain 1.5 8B \cite{physbrain1.5}.

\begin{table*}[t]
    \centering
    \caption{\textbf{Results on the embodied benchmark.}
    Average values follow the corresponding evaluation runs. Dashes denote
    unavailable results. The best result on each benchmark is boldfaced,
    including ties.}
    \label{tab:cognitive_embodied}
    \renewcommand{\arraystretch}{1.14}
    \setlength{\tabcolsep}{3.2pt}
    \scriptsize
    \begin{adjustbox}{max width=\textwidth}
    \begin{tabular}{@{}llcccccc@{}}
        \toprule
        Subcategory & Benchmark & \makecell{ME-VLM\\4B} &
        \makecell{ME-VLM\\35B-A3B} & \makecell{Hy-Embodied\\30B-A3B} &
        \makecell{RynnBrain1.1\\9B} & \makecell{Mimo-Embodied\\7B} &
        \makecell{PhysBrain 1.5\\8B} \\
        \midrule
        & \textbf{Average} & 63.4 & \textbf{70.9} & 60.2 & 62.1 & 54.3 & 62.7 \\
        \midrule
        \multirow{15}{*}{\makecell[l]{Physical\\Understanding}}
        & CV-Bench & 89.5 & \textbf{91.7} & 89.7 & 88.2 & 88.8 & 90.0 \\
        & VABench-Point & 67.0 & \textbf{75.4} & 59.7 & 20.3 & 33.3 & 65.2 \\
        & VABench-Visual & 81.6 & 82.6 & 79.7 & 87.6 & 66.9 & \textbf{89.8} \\
        & RefCOCO-testA & 85.9 & 87.8 & 49.4 & \textbf{89.5} & 74.5 & 60.9 \\
        & RefCOCO-testB & 76.6 & 80.3 & 56.2 & \textbf{81.6} & 67.4 & 62.8 \\
        & ERQA & 45.8 & 54.3 & \textbf{60.8} & 47.5 & 46.8 & 52.8 \\
        & MindCube & -- & \textbf{91.1} & 70.0 & 86.9 & 35.2 & 86.2 \\
        & SPARBench & 57.3 & \textbf{67.6} & 53.4 & 51.1 & 41.2 & 53.8 \\
        & SPBench-MV & 69.9 & \textbf{83.5} & 59.8 & 74.4 & 49.4 & 78.8 \\
        & SPBench-SI & 70.6 & 71.7 & 50.5 & 59.0 & 44.8 & \textbf{77.7} \\
        & MMSI-Bench & 33.3 & 46.4 & 41.8 & \textbf{47.0} & 29.6 & 41.0 \\
        & ViewSpatial-Bench & 57.8 & \textbf{62.7} & 53.3 & 54.2 & 40.5 & 62.5 \\
        & EmbSpatial-Bench & 81.0 & \textbf{83.1} & 82.7 & 81.9 & 76.2 & 81.8 \\
        & RefSpatial-Bench & 46.9 & 63.4 & 53.4 & \textbf{67.2} & 48.0 & 50.9 \\
        & RoboSpatial-Home & 67.7 & \textbf{74.0} & 69.4 & 69.1 & 61.8 & 73.9 \\
        \midrule
        \multirow{3}{*}{\makecell[l]{Task\\Planning}}
        & Cosmos & 65.7 & \textbf{76.9} & 66.9 & 56.1 & 56.8 & 72.8 \\
        & EgoPlan2 & 55.3 & \textbf{64.5} & 49.6 & 43.5 & 43.0 & 62.1 \\
        & RoboBench-Planning & 40.1 & 53.7 & 54.9 & \textbf{59.8} & 57.2 & 42.2 \\
        \midrule
        \multirow{4}{*}{\makecell[l]{Action\\Execution}}
        & VSIBench & 63.2 & 71.2 & 58.9 & \textbf{74.9} & 48.5 & 61.9 \\
        & SITE-Bench-Video & 65.6 & \textbf{72.9} & 69.2 & 68.9 & 59.0 & 67.5 \\
        & RoboBench-Perception & 41.8 & 52.2 & \textbf{55.9} & 40.5 & 34.9 & 31.8 \\
        & RoboBench-Affordance & 47.0 & 60.7 & \textbf{61.7} & 33.0 & 36.7 & 25.6 \\
        \midrule
        \multirow{4}{*}{Correction}
        & RoboFail-Execution & 79.7 & \textbf{86.3} & 62.8 & 77.1 & 72.6 & 77.8 \\
        & RoboFail-Planning & 56.7 & \textbf{63.3} & 53.3 & \textbf{63.3} & \textbf{63.3} & 56.7 \\
        & RoboFAC & 74.7 & \textbf{75.7} & 51.0 & 55.4 & 61.2 & 64.8 \\
        & RoboBench-Error & -- & 52.7 & \textbf{53.0} & 42.2 & 33.5 & 34.1 \\
        \bottomrule
    \end{tabular}
    \end{adjustbox}
\end{table*}

ME-VLM 35B-A3B achieves the highest overall average of 70.9, exceeding the
strongest comparison model by 8.2 points and obtaining the best or tied-best
result on 14 of the 26 benchmarks. Its advantages are concentrated in the
capabilities that connect physical understanding to executable behavior. The
model reaches 91.1 on MindCube, 83.5 on SPBench-MV, 83.1 on
EmbSpatial-Bench, and 74.0 on RoboSpatial-Home, establishing strong spatial
reasoning across complementary task formats. It also leads on Cosmos and
EgoPlan2 with 76.9 and 64.5, and obtains 72.9 on SITE-Bench-Video.

The execution-correction results further validate the closed-loop cognitive
design. ME-VLM achieves 86.3 on RoboFail-Execution, 63.3 on
RoboFail-Planning, and 75.7 on RoboFAC. Together with the broad gains in
physical and spatial understanding, these results demonstrate that Cognitive
Core connects perception, planning, execution understanding, and recovery in a
single model rather than optimizing an isolated embodied skill.

\subsubsection{Agent Benchmark}

The agent evaluation covers multimodal understanding, tool and skill
invocation, long-horizon planning, reasoning, and instruction following. We
compare ME-VLM at 35B-A3B and 4B scales with Hy-Embodied 30B-A3B,
RynnBrain1.1 9B, Mimo-Embodied 7B, and PhysBrain 1.5 8B.

\begin{table*}[t]
    \centering
    \caption{\textbf{Results on the agent benchmark.}
    The average follows the benchmark suite shown in the table. Dashes denote
    unavailable results. The best result on each benchmark is boldfaced,
    including ties.}
    \label{tab:cognitive_agent}
    \renewcommand{\arraystretch}{1.14}
    \setlength{\tabcolsep}{3.2pt}
    \scriptsize
    \begin{adjustbox}{max width=\textwidth}
    \begin{tabular}{@{}llcccccc@{}}
        \toprule
        Subcategory & Benchmark & \makecell{ME-VLM\\4B} &
        \makecell{ME-VLM\\35B-A3B} & \makecell{Hy-Embodied\\30B-A3B} &
        \makecell{RynnBrain1.1\\9B} & \makecell{Mimo-Embodied\\7B} &
        \makecell{PhysBrain 1.5\\8B} \\
        \midrule
        & \textbf{Average} & 63.1 & \textbf{72.5} & 61.2 & 62.9 & 54.1 & 46.6 \\
        \midrule
        \multirow{5}{*}{\makecell[l]{Multimodal\\Understanding}}
        & BLINK & 58.9 & \textbf{67.7} & 66.2 & 53.8 & 56.3 & 64.7 \\
        & MMStar & 72.5 & \textbf{77.3} & 76.1 & 68.7 & 68.7 & 62.5 \\
        & MVBench & 64.1 & \textbf{71.9} & 67.6 & 62.3 & 56.7 & 60.8 \\
        & VideoMME & 55.6 & \textbf{60.3} & 58.2 & 55.4 & 25.1 & 52.2 \\
        & RealWorldQA & 79.0 & \textbf{79.7} & 76.6 & 39.5 & 67.1 & 71.5 \\
        \midrule
        \multirow{2}{*}{\makecell[l]{Tool or Skill\\Invocation}}
        & BFCL-V4 & 40.5 & \textbf{63.2} & -- & \textbf{63.2} & 51.5 & -- \\
        & TAU2-Bench & 50.1 & 63.9 & -- & \textbf{75.6} & 15.0 & -- \\
        \midrule
        \multirow{2}{*}{\makecell[l]{Long-Horizon\\Planning}}
        & ClawEval avg3 & -- & 58.0 & -- & \textbf{62.4} & 37.0 & -- \\
        & ClawEval pass$^3$ & -- & 31.8 & -- & \textbf{37.2} & 6.0 & -- \\
        \midrule
        \multirow{5}{*}{Reasoning}
        & GPQA & 67.2 & \textbf{78.7} & 46.5 & 75.3 & 54.9 & 35.0 \\
        & AIME25 & 47.5 & \textbf{72.9} & 26.7 & 70.4 & 48.3 & 3.8 \\
        & MMLU-Pro & 75.4 & \textbf{82.8} & 74.3 & 81.3 & 71.6 & 46.7 \\
        & MMMU-Pro & 58.4 & \textbf{69.3} & 64.8 & 60.8 & 52.1 & 35.1 \\
        & MathVision & 59.6 & \textbf{74.3} & 66.1 & 61.4 & 54.5 & 19.6 \\
        \midrule
        \multirow{2}{*}{\makecell[l]{Instruction\\Following}}
        & IFEval & 81.9 & \textbf{86.9} & 82.1 & 83.6 & 65.6 & 79.1 \\
        & IFBench & 37.4 & \textbf{48.3} & 29.3 & 42.5 & 27.9 & 27.6 \\
        \bottomrule
    \end{tabular}
    \end{adjustbox}
\end{table*}

ME-VLM 35B-A3B achieves the highest overall average of 72.5, outperforming the
strongest comparison model by 9.6 points. The 4B model also reaches 63.1,
surpassing every comparison model in overall average. The 35B-A3B model ranks
first on all five multimodal
understanding benchmarks and all five reasoning benchmarks, demonstrating that
embodied specialization preserves a broad multimodal and reasoning foundation.
The model reaches 71.9 on MVBench, 60.3 on VideoMME, 78.7 on GPQA, 72.9 on
AIME25, and 74.3 on MathVision.

The same unified model also delivers strong structured interaction and
instruction execution. It matches the best result on BFCL-V4 at 63.2 and
achieves 86.9 on IFEval and 48.3 on IFBench, leading the strongest comparison
model by 3.3 and 5.8 points, respectively. These results establish Cognitive
Core as a balanced foundation that combines physical-world cognition with the
multimodal reasoning, tool use, and instruction-following capabilities required
for experience evolution.

\subsection{Evaluation of Action Model}
\label{sec:action_evaluation}

We evaluate ME-Brain on history-dependent manipulation, general manipulation
capabilities, and real-world robot deployment. RoboMME tests whether the model
can use past visual evidence to guide current actions. RoboDojo evaluates
generalization, precision, long-horizon execution, memory, and open-vocabulary
instruction following. Experiments on a physical Piper dual-arm robot further
assess execution performance during real-world interaction.

\subsubsection{Memory Benchmark}
\label{sec:action_memory_benchmark}

We first evaluate the role of historical information in action generation on
RoboMME~\citep{dai2026robomme}. The benchmark contains 16 tasks organized into
four suites according to their memory requirements: Counting, Permanence,
Reference, and Imitation, with four tasks per suite.
Table~\ref{tab:action_robomme} reports the arithmetic mean of task success rates
within each suite and the equally weighted mean across the four suites.
We compare ME-Brain with $\pi_{0.5}$, FrameSampling + Modulation, and
MemER~\citep{sridhar2025memer}.

\begin{table}[tbp]
    \centering
    \caption{\textbf{Success rates on RoboMME (\%).}
    AVG is the equally weighted mean across the four suites.}
    \label{tab:action_robomme}
    \small
    \setlength{\tabcolsep}{4pt}
    \begin{tabular*}{\linewidth}{@{\extracolsep{\fill}}lrrrrr@{}}
        \toprule
        Method & Counting & Permanence & Reference & Imitation & AVG \\
        \midrule
        $\pi_{0.5}$ & 28.0 & 19.5 & 16.0 & 9.0 & 18.12 \\
        FrameSampling + Modulation & 66.0 & 24.0 & 36.5 & 52.0 & 44.62 \\
        MemER & 48.5 & 51.5 & 42.0 & 25.0 & 41.75 \\
        \textbf{ME-Brain} & \textbf{70.0} & 27.5 & 39.0 & \textbf{55.0} & \textbf{47.88} \\
        \bottomrule
    \end{tabular*}
\end{table}

\begin{table}[!t]
    \centering
    \caption{\textbf{Results on the RoboDojo simulation benchmark.}
    Comparison results are taken from the
    \href{https://robodojo-benchmark.com/leaderboard}{official RoboDojo leaderboard}.
    Score ranges from 0 to 100, and SR is reported in percent.
    Generalization aggregates Gen-Std and Gen-Rand, and AVERAGE equally weights
    the five capability dimensions. Best results are boldfaced, and the
    ME-Brain rows are shaded.}
    \label{tab:action_robodojo}
    \scriptsize
    \setlength{\tabcolsep}{2.8pt}
    \begin{adjustbox}{max width=\linewidth}
    \begin{tabular}{@{}llrrrrrr@{}}
        \toprule
        Model & Metric & AVERAGE & Generalization & Precision & Long-Horizon & Memory & Open \\
        \midrule
        \multirow{2}{*}{GalaxeaVLA (G0.5)} & Score & 20.23 & 18.46 & \textbf{28.25} & \textbf{44.12} & 8.61 & 1.73 \\
        & SR (\%) & 14.88 & 12.83 & \textbf{20.42} & \textbf{32.25} & 7.33 & 1.58 \\
        \addlinespace[2pt]
        \multirow{2}{*}{Xiaomi-Robotics-1} & Score & 20.07 & \textbf{23.54} & 26.69 & 38.39 & 7.81 & \textbf{3.94} \\
        & SR (\%) & 13.93 & \textbf{17.00} & 18.83 & 23.67 & 6.56 & \textbf{3.58} \\
        \addlinespace[2pt]
        \multirow{2}{*}{OpenWAM-$\alpha$} & Score & 17.18 & 20.71 & 18.45 & 34.93 & 10.41 & 1.41 \\
        & SR (\%) & 11.92 & 14.83 & 9.25 & 25.33 & 9.11 & 1.08 \\
        \addlinespace[2pt]
        \multirow{2}{*}{$\pi_{0.5}$} & Score & 11.41 & 13.38 & 12.40 & 23.54 & 5.78 & 1.98 \\
        & SR (\%) & 6.91 & 8.17 & 5.50 & 14.67 & 4.56 & 1.67 \\
        \midrule
        \textbf{ME-Brain} & Score & \textbf{21.51} & 21.95 & 22.04 & 34.55 & \textbf{25.57} & 3.45 \\
        & SR (\%) & \textbf{16.03} & 16.00 & 12.75 & 24.00 & \textbf{24.67} & 2.75 \\
        \bottomrule
    \end{tabular}
    \end{adjustbox}
\end{table}

ME-Brain achieves an overall success rate of $47.88\%$, improving over the
in-house $\pi_{0.5}$ baseline, FrameSampling + Modulation, and MemER by
approximately 29.8, 3.3, and 6.1 percentage points, respectively.
Compared with the baseline that combines historical-frame sampling and
modulation, ME-Brain improves performance on all four suites. In particular,
it reaches $70.0\%$ on Counting and $55.0\%$ on Imitation, indicating that event
counts and historical action patterns can inform current manipulation decisions.
The consistent gains across these memory requirements suggest that
action-conditioned history retrieval and event-driven action generation work
effectively together.

\subsubsection{Action Benchmark}
\label{sec:action_benchmark}

We further evaluate general manipulation capabilities on the RoboDojo simulation
benchmark~\citep{chen2026robodojo}. The evaluation covers five capability
dimensions on the ARX X5 dual-arm platform. Generalization is tested in both
standard scenes (Gen-Std) and randomized scenes (Gen-Rand). Each of the 54 task
configurations is evaluated over 24 trials, yielding 1296 valid trials in total.
We report both terminal success rate (SR) and Score, a measure of task completion
progress. Following the benchmark's aggregation scheme, Generalization combines
Gen-Std and Gen-Rand, while AVERAGE equally weights Generalization, Precision,
Long-Horizon, Memory, and Open.

\begin{table}[!t]
    \centering
    \caption{\textbf{Evaluation on the six-task ME-RealBench using a Piper dual-arm robot.}
    (a) Overall results: SR and Score are equally weighted means across tasks;
    TimeExpend is the mean execution time over successful trials only.
    (b) Per-task results: SR is reported in percent and Score ranges from 0 to 100,
    with 10 trials per task. Higher SR and Score and lower TimeExpend are better.
    Best values for each metric are boldfaced, including ties.}
    \label{tab:action_piper}
    \small
    \setlength{\tabcolsep}{4pt}
    \textbf{(a) Overall performance}\par\smallskip
    \begin{tabular*}{0.9\linewidth}{@{\extracolsep{\fill}}lrrr@{}}
        \toprule
        Model & SR (\%) $\uparrow$ & Score $\uparrow$ & TimeExpend (s) $\downarrow$ \\
        \midrule
        \textbf{ME-Brain} & \textbf{66.7} & \textbf{69.5} & 75.5 \\
        DM0.5 & 55.0 & 56.7 & 89.1 \\
        $\pi_{0.5}$ & 50.0 & 52.2 & \textbf{67.5} \\
        xr-1 & 23.3 & 29.8 & 77.4 \\
        \bottomrule
    \end{tabular*}

    \medskip
    \textbf{(b) Per-task performance}\par\smallskip
    \begin{tabular*}{\linewidth}{@{\extracolsep{\fill}}llrrrrrr@{}}
        \toprule
        Model & Metric & \shortstack{Stack\\bowls} & \shortstack{Cover\\blocks}
            & \shortstack{Stand\\bottle} & \shortstack{Place in\\basket}
            & \shortstack{Fill pen\\holder} & \shortstack{Plug in\\charger} \\
        \midrule
        \multirow{2}{*}{\textbf{ME-Brain}} & SR (\%) & \textbf{100} & \textbf{50} & \textbf{90} & 90 & 60 & \textbf{10} \\
            & Score & \textbf{100} & \textbf{50} & \textbf{95} & 90 & 72 & \textbf{10} \\
        \addlinespace[2pt]
        \multirow{2}{*}{DM0.5} & SR (\%) & \textbf{100} & 10 & 80 & 60 & \textbf{80} & 0 \\
            & Score & \textbf{100} & 10 & 84 & 60 & \textbf{86} & 0 \\
        \addlinespace[2pt]
        \multirow{2}{*}{$\pi_{0.5}$} & SR (\%) & \textbf{100} & 0 & \textbf{90} & \textbf{100} & 0 & \textbf{10} \\
            & Score & \textbf{100} & 0 & \textbf{95} & \textbf{100} & 8 & \textbf{10} \\
        \addlinespace[2pt]
        \multirow{2}{*}{xr-1} & SR (\%) & 70 & 0 & 50 & 20 & 0 & 0 \\
            & Score & 73 & 15 & 61 & 20 & 10 & 0 \\
        \bottomrule
    \end{tabular*}
\end{table}

As shown in Table~\ref{tab:action_robodojo}, ME-Brain achieves an average Score
of 21.51 and an average SR of $16.03\%$, exceeding the official $\pi_{0.5}$
baseline by 10.10 points and 9.12 percentage points, respectively. ME-Brain
improves both metrics across all five capability dimensions. The largest gain
appears in Memory, where Score rises from 5.78 to 25.57 and SR from $4.56\%$ to
$24.67\%$. It also improves Long-Horizon Score by 11.01 points and SR by 9.33
percentage points, while Precision gains 9.64 points in Score and 7.25
percentage points in SR. Among the five systems in
Table~\ref{tab:action_robodojo}, ME-Brain ranks first overall in both Score and
SR, leading the second-best system by 1.28 points and 1.15 percentage points,
respectively. It also achieves top-two performance on Generalization, Memory,
and Open for both metrics. In particular, ME-Brain ranks first on Memory with a
Score of 25.57 and an SR of $24.67\%$, exceeding the second-best results by
15.16 points and 15.56 percentage points. These results demonstrate that
event-target guidance, local future modeling, and action-conditioned memory
modulation form a balanced policy with distinctive strength in memory-dependent
control.

\subsection{Real-World Robot Evaluation}
\label{sec:action_real_world}

We introduce ME-RealBench to systematically evaluate ME-Brain's task execution
capabilities in real-world physical environments. Deployed on a Piper dual-arm
robot, the benchmark comprises six tasks: stacking bowls, stacking blocks and
covering them with a cup, standing a bottle upright, placing objects in a basket,
inserting pens into a pen holder, and plugging in a charger. These tasks cover
key capabilities including dual-arm coordination, fine-grained manipulation,
and multi-step task execution. Each task is evaluated over 10 independent trials.
Success Rate (SR) is computed from the binary terminal success outcome of each
trial. Partially completed trajectories receive a Score between 0 and 100 based
on RoboDojo's task-specific scoring levels, capturing task progress even when
the task is not fully completed. Three evaluators independently score each trial
under a double-blind protocol. Their ratings are averaged, followed by averaging
over trials within each task and then equally over tasks. TimeExpend is measured
from the policy's first action output until task success, failure, timeout, or a
safety stop, and is averaged over successful trials only.

ME-Brain achieves a mean SR of $66.7\%$ and a mean Score of 69.5 across the six
tasks, the highest values among the four evaluated models
(Table~\ref{tab:action_piper}). Relative to DM0.5, these metrics improve by
11.7 percentage points and 12.8 points, respectively; relative to $\pi_{0.5}$,
the improvements are 16.7 percentage points and 17.3 points. ME-Brain obtains a
non-zero Score on every task and succeeds in all 10 bowl-stacking trials. On the
more challenging block-covering task, it reaches a $50\%$ SR, 40 percentage
points above DM0.5. On the pen-holder task, which requires repeated grasping and
placing, its Score reaches 72, improving over the $\pi_{0.5}$ Score of 8 by 64
points. Together, the simulation and real-world results demonstrate effective
execution across diverse manipulation tasks for the interaction-event-centric
policy.

%% file: sec/08_Qualitative_Examples.tex
% \clearpage
\section{Qualitative Examples}
\label{sec:qualitative-examples}

\begin{figure}[!t]
    \centering
    \includegraphics[width=\linewidth,height=0.58\textheight,keepaspectratio]{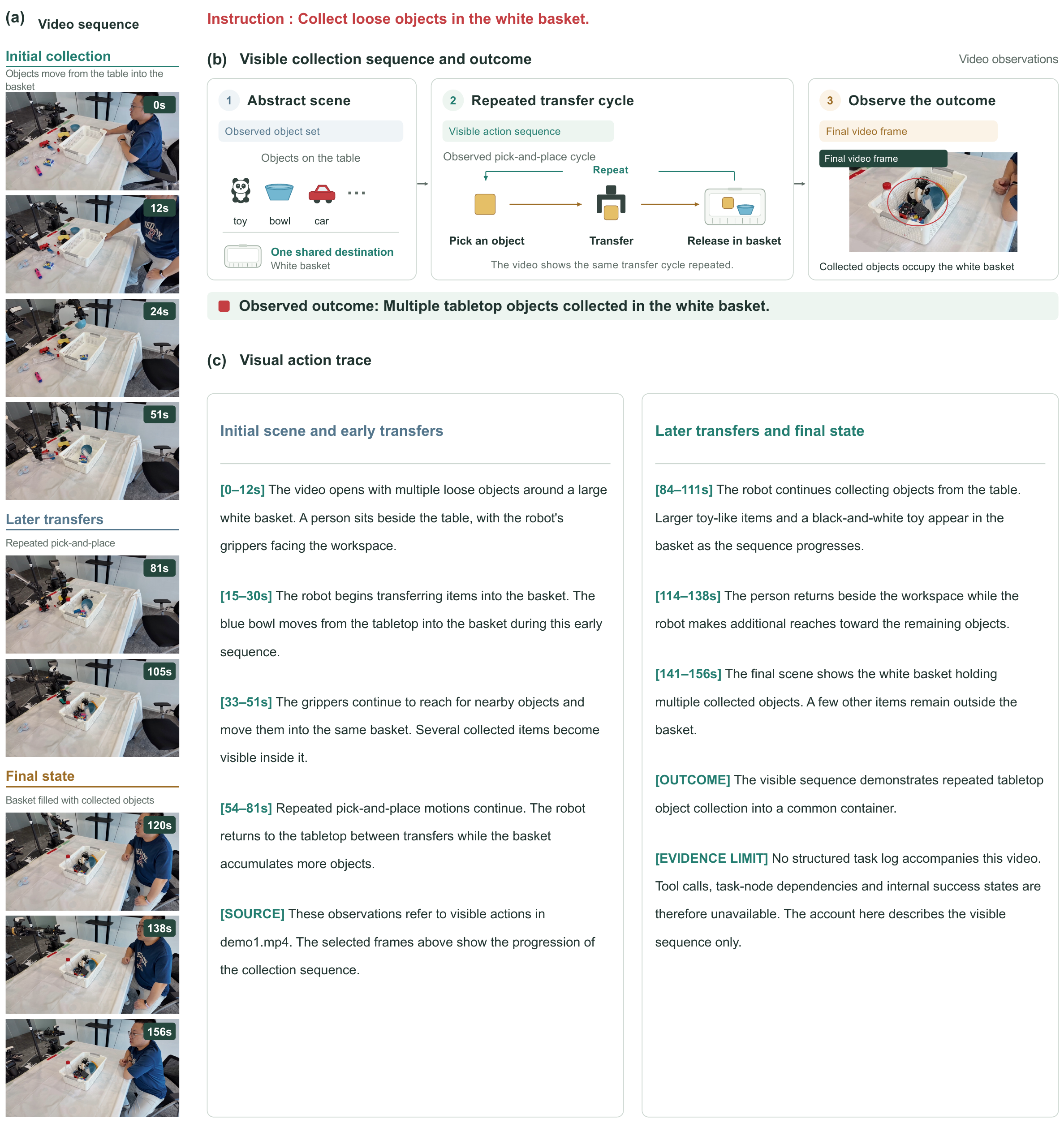}
    \caption{Multi-object tabletop collection}
    \label{fig:qualitative-case1}
\end{figure}

This section presents six real-world deployment cases that qualitatively illustrate how the three core modules of ME-Brain work together. Spanning capabilities from basic manipulation to memory-guided reasoning, these cases show how the system integrates Evolvable Memory, Cognitive Core, and Action Model into coherent task execution in physical environments.

The first case illustrates ME-Brain's ability to sustain execution over repeated manipulation cycles. The robot repeatedly grasps, transfers, and releases objects scattered across the tabletop into the same white basket. As the target objects and their positions change throughout execution, the robot returns to the tabletop between transfers to locate and grasp additional objects, progressively completing the collection process. The key frames show the initial scene, intermediate object transfers, and the growing number of objects in the basket, illustrating the system's ability to maintain a repeated action pattern during continuous manipulation. This analysis is based primarily on actions visible in the video and documents the ongoing collection of multiple objects.

\begin{figure}[!t]
    \centering
    \includegraphics[width=\linewidth,height=0.63\textheight,keepaspectratio]{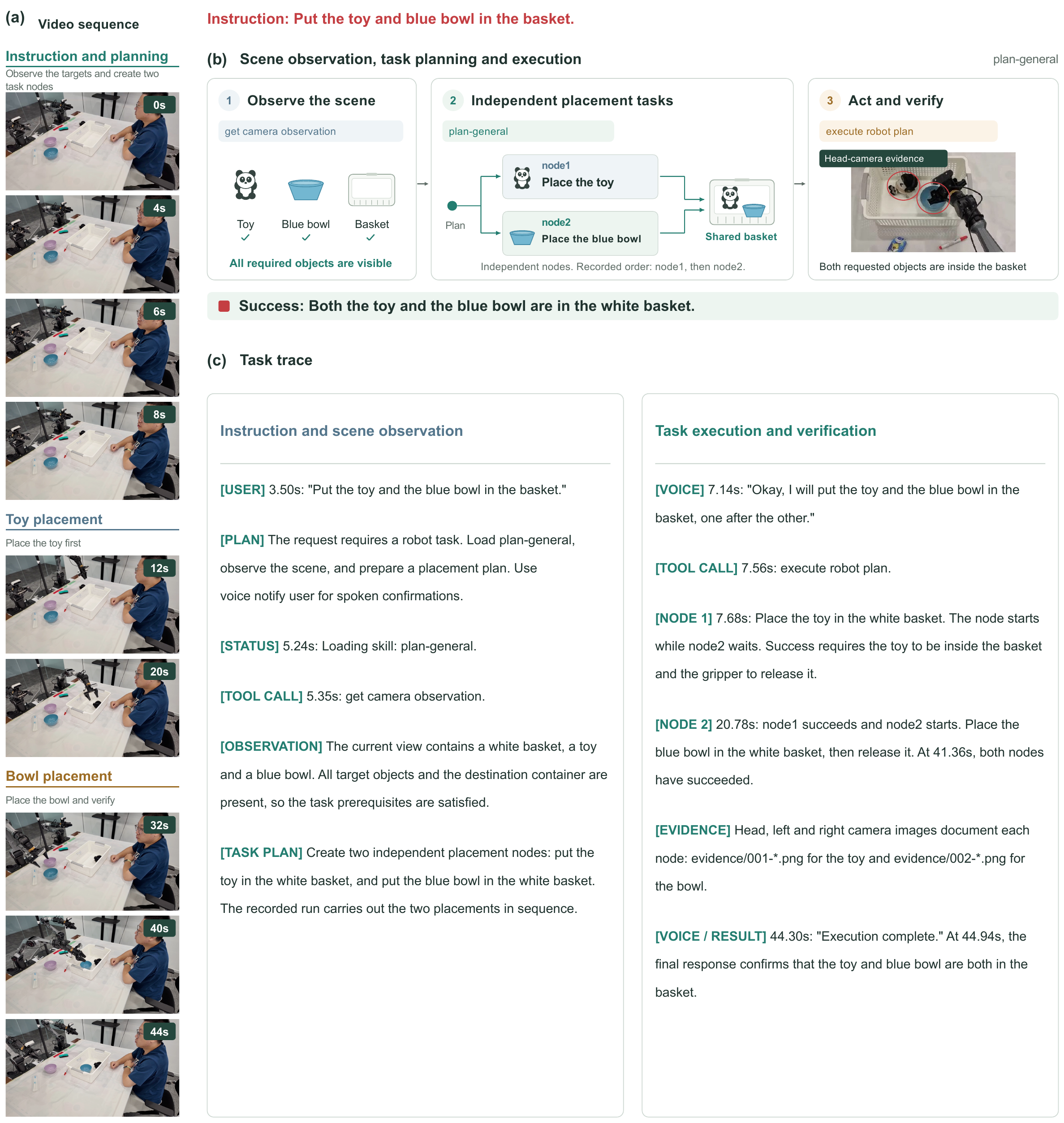}
    \caption{Instruction-guided object placement}
    \label{fig:qualitative-case2}
\end{figure}

The second case illustrates ME-Brain's ability to translate a language instruction with multiple goals into structured subtasks and execute them sequentially. Given the instruction ``Put the toy and the blue bowl in the basket,'' the system first uses current visual observations to locate both target objects and the white basket, then creates two task nodes for placing the toy and the blue bowl. The two subgoals are logically independent. In this run, the system first places the toy in the basket and releases the gripper, then starts the blue-bowl placement, ultimately completing both subgoals. The task trace records scene observation, task planning, node execution, and completion feedback, illustrating how ME-Brain extracts target objects and their placement relations from language and translates them into a sequence of robot actions.

\begin{figure}[!t]
    \centering
    \includegraphics[width=\linewidth,height=0.63\textheight,keepaspectratio]{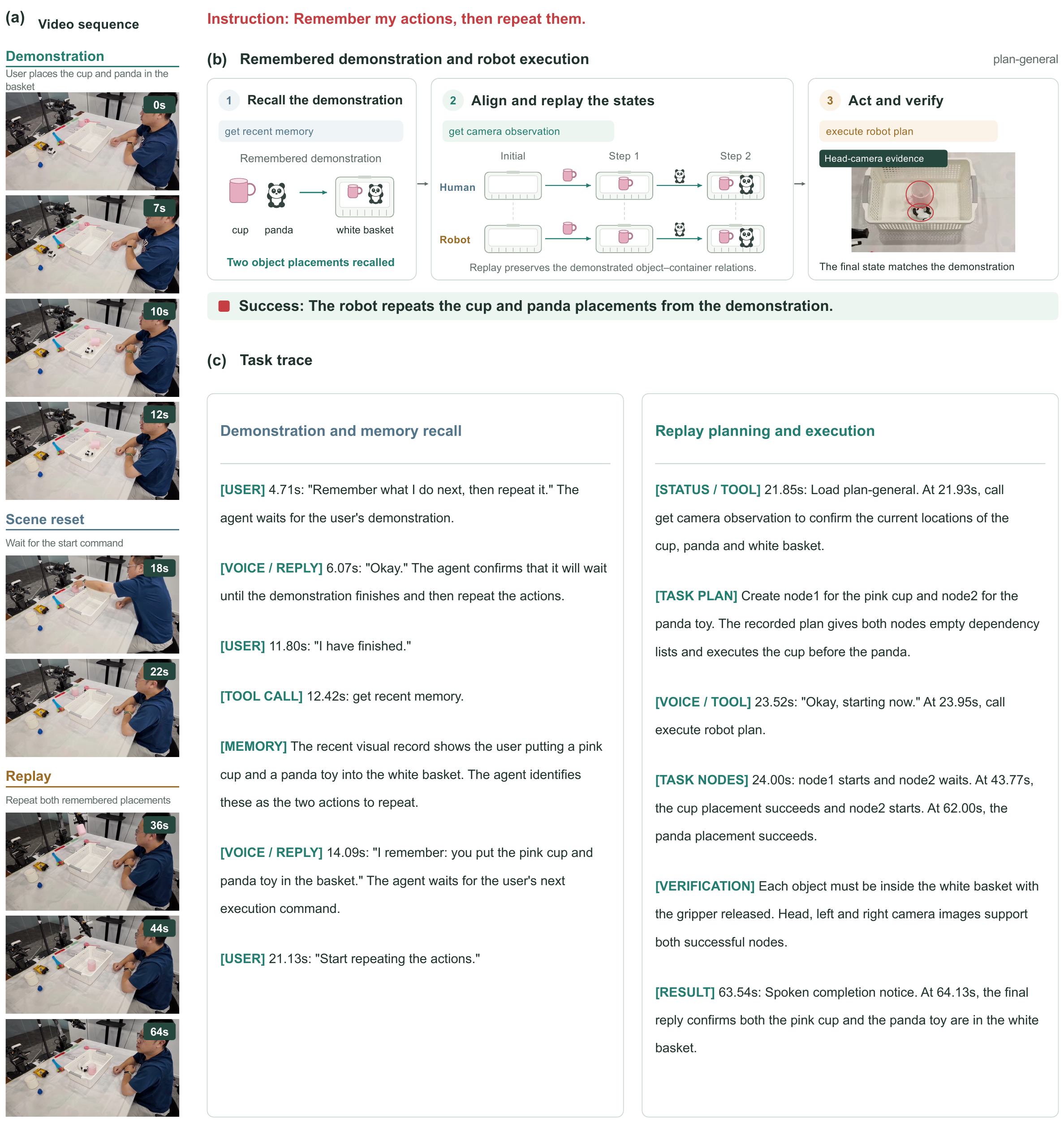}
    \caption{Visual memory-guided demonstration replay}
    \label{fig:qualitative-case3}
\end{figure}

The third case illustrates ME-Brain's ability to reproduce a human manipulation demonstration using recent visual memory. After receiving the instruction ``Remember my actions, then repeat them,'' the system waits for the user to complete the demonstration. The user places a pink cup and a panda toy into the white basket and announces that the demonstration is complete. The system then retrieves recent visual memory to identify and confirm the two placement actions to be repeated. After the scene is reset and the objects are returned to the tabletop, the user gives the start command. The system confirms the current locations of the objects and basket, then places the pink cup and the panda toy into the basket in sequence. This case reproduces both the object--container relations and the placement order in the demonstration, illustrating how recent visual memory preserves demonstration content and supports task execution by combining historical information with current scene observations.

\clearpage
\begin{figure}[!t]
    \centering
    \includegraphics[width=\linewidth,height=0.63\textheight,keepaspectratio]{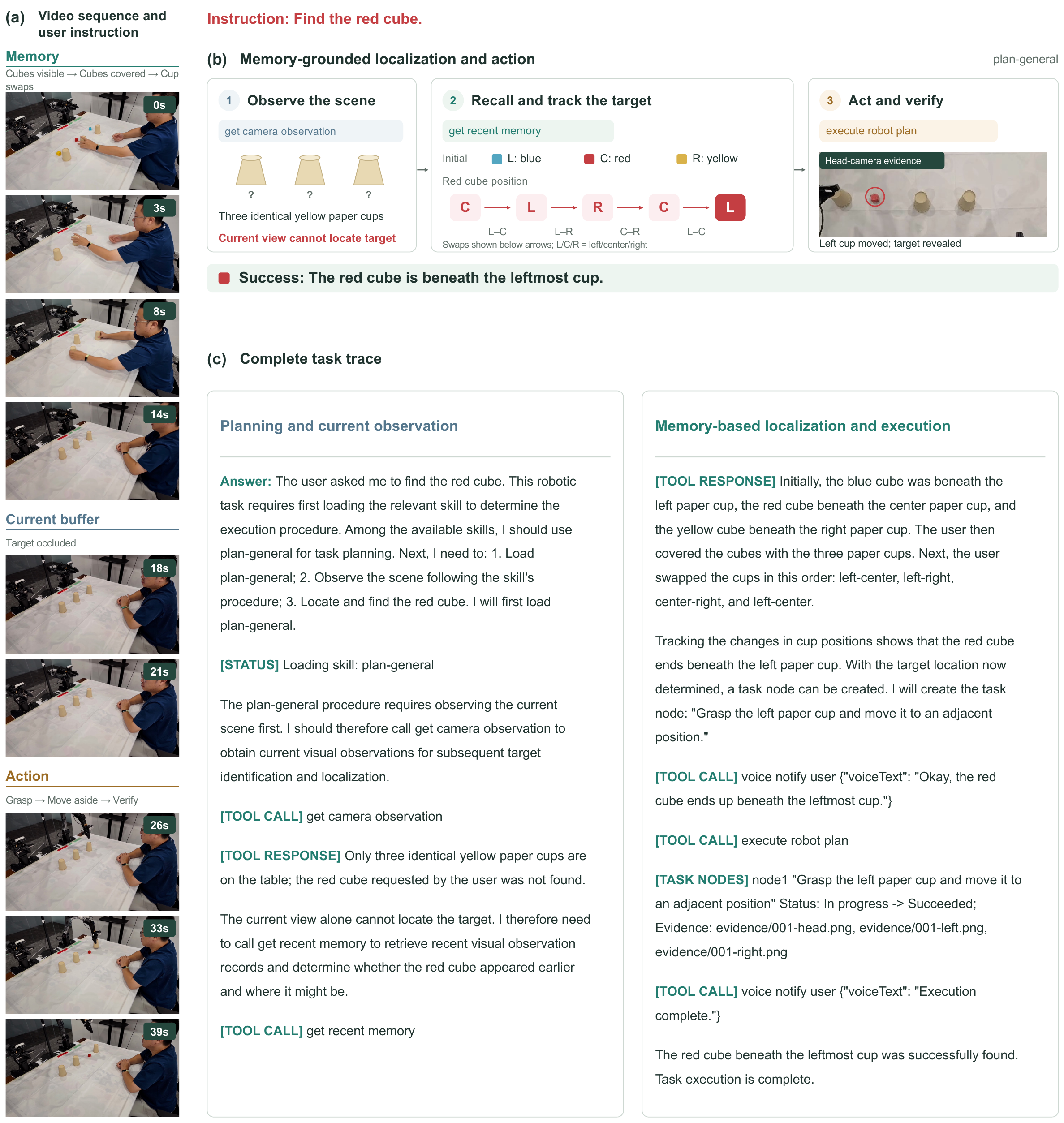}
    \caption{Visual memory-guided localization and action}
    \label{fig:qualitative-case4}
\end{figure}

The fourth case illustrates ME-Brain's ability to locate an occluded object using a sequence of past events. Given the instruction ``Find the red cube,'' the current observation shows only three identical yellow paper cups, making it impossible to determine the target's location directly. By retrieving recent visual memory, the system establishes that the red cube was initially beneath the center cup and reconstructs four subsequent cup swaps: left--center, left--right, center--right, and left--center. Tracking these position changes step by step, the system infers that the red cube is now beneath the leftmost cup and commands the robot to move that cup aside, revealing the target. Whereas the third case reproduces demonstrated placement relations, this case requires the system to integrate state changes across multiple time steps, infer the current location of an invisible target from past events, and verify the inference through physical action.

\clearpage
\begin{figure}[!t]
    \centering
    \includegraphics[width=\linewidth,height=0.63\textheight,keepaspectratio]{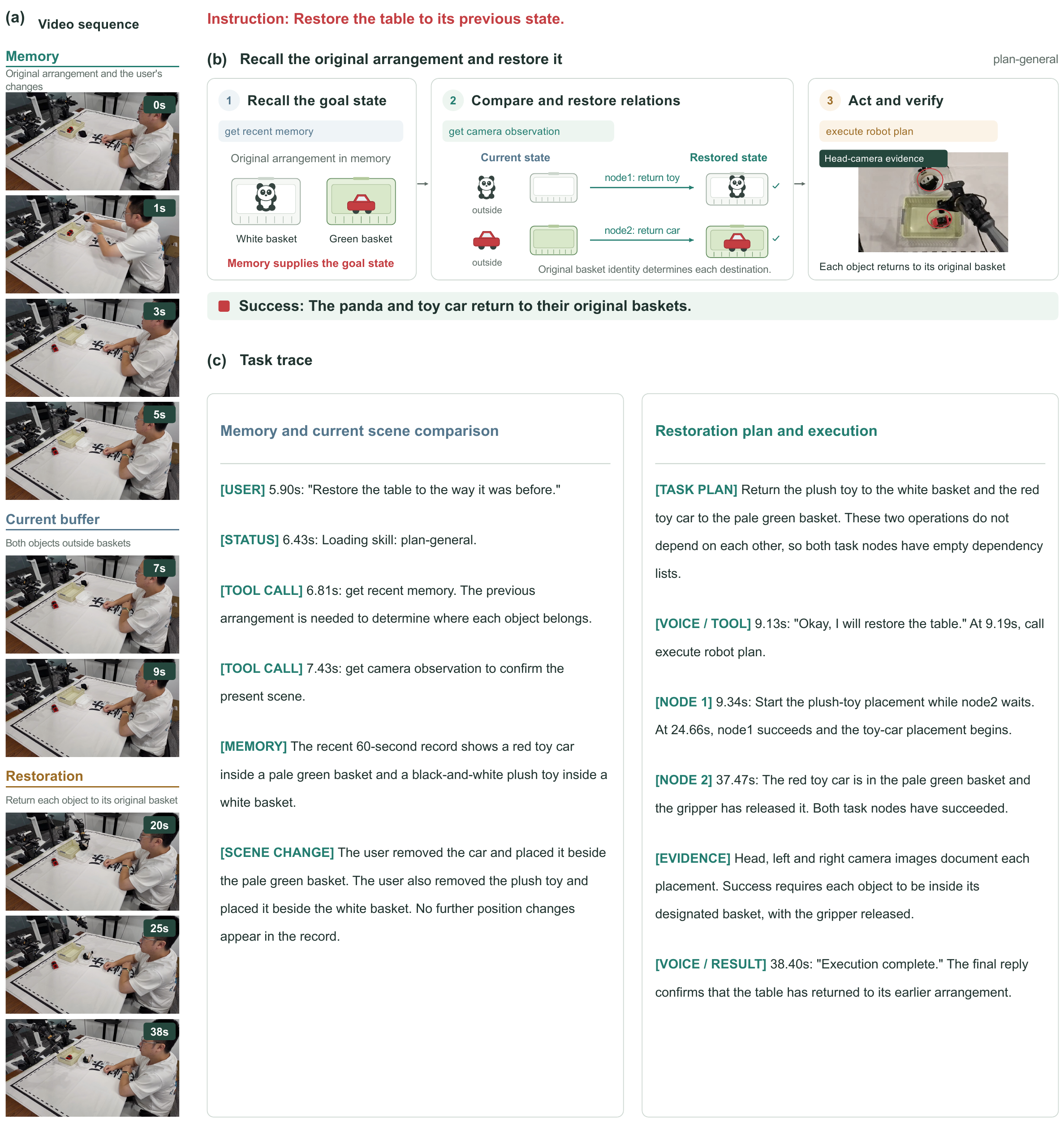}
    \caption{Visual memory-guided scene restoration}
    \label{fig:qualitative-case5}
\end{figure}

The fifth case illustrates ME-Brain's ability to recover a task's goal state from a past scene. After removing objects from their original containers, the user instructs the system to ``Restore the table to its previous state'' without specifying a destination for each object. The system first retrieves recent visual memory to establish that the panda toy was originally in the white basket and the red toy car in the pale green basket. It then combines this information with current observations to determine the required restoration actions. The robot returns the panda toy to the white basket, followed by the red toy car to the pale green basket, restoring the original object--container relations. Compared with demonstration replay and occluded-object localization, this case emphasizes retrieving and restoring a historical scene state: the system must recover goal relations left unspecified by the current instruction from memory and use them to generate a concrete execution plan.

\clearpage
\begin{figure}[!t]
    \centering
    \includegraphics[width=\linewidth,height=0.63\textheight,keepaspectratio]{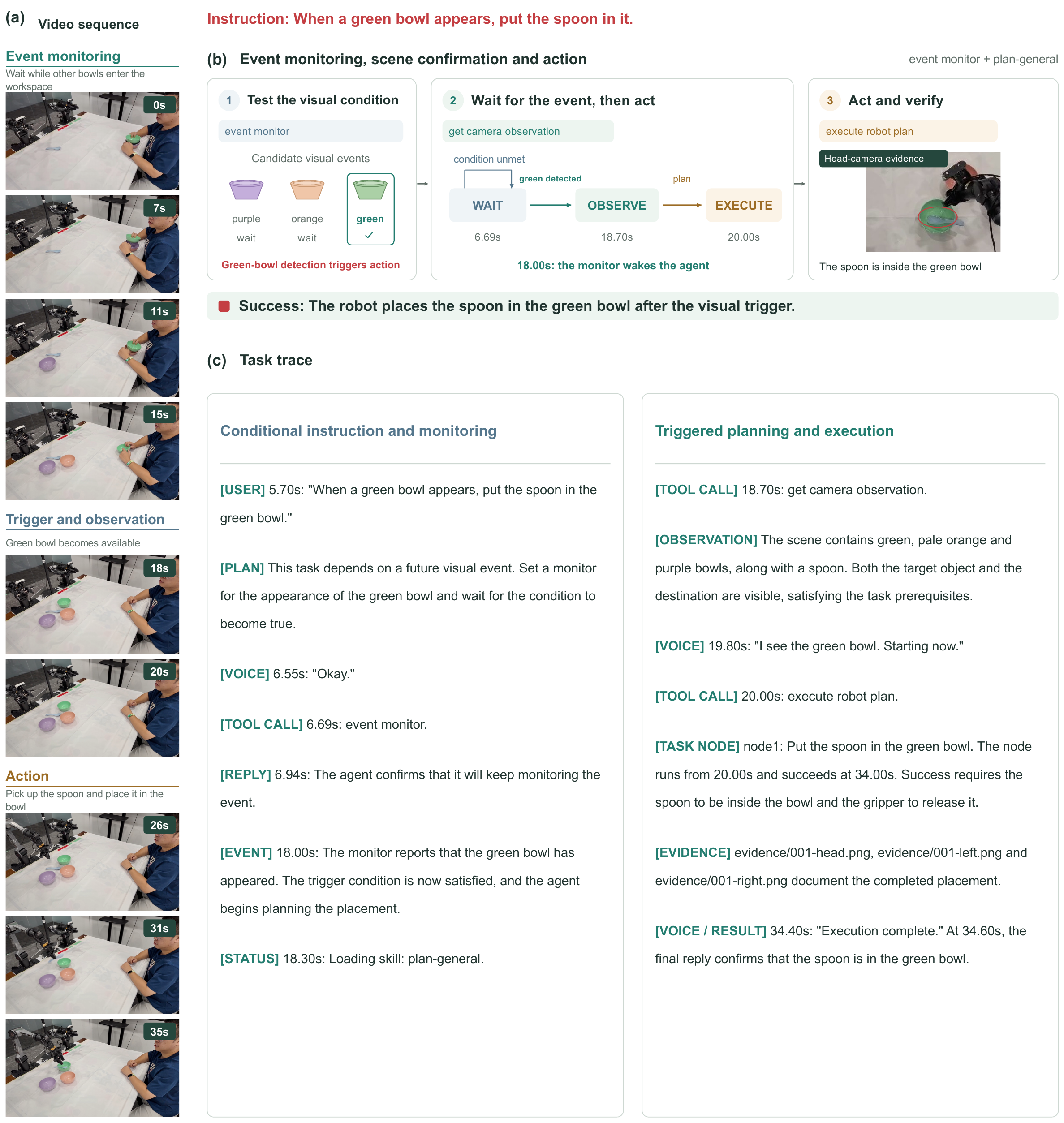}
    \caption{Visual event-triggered object placement}
    \label{fig:qualitative-case6}
\end{figure}

The sixth case illustrates ME-Brain's ability to interpret conditional language instructions and execute actions in response to an event. Given the instruction ``When a green bowl appears, put the spoon in it,'' the system sets the appearance of a green bowl as the condition for visual event monitoring and enters a waiting state. It continues waiting as a purple bowl and an orange bowl appear in succession, since neither satisfies the target condition. Once the monitor reports the appearance of the green bowl, the system obtains current visual observations, confirms the locations of the spoon and the green bowl, and plans and executes the placement task, finally placing the spoon in the bowl and releasing the gripper. In this case, the user specifies the task in advance, while a subsequent environmental event determines when execution begins. The case illustrates how ME-Brain converts a conditional language instruction into a monitoring condition and, once that condition is met, connects scene confirmation, task planning, and robot execution.
\clearpage

%% file: sec/09_Conclusion_and_Future_Works.tex
\section{Conclusion and Future Work}

\subsection{Conclusion}
This work addresses the limitation that embodied models largely remain fixed
after training and cannot continuously learn from ongoing interaction. We
propose \textit{ME-Brain}, a self-evolving embodied system that decomposes
post-deployment self-evolution into three interconnected problems: experience
acquisition, experience evolution, and action execution. These problems are
respectively handled by \textit{Evolvable Memory}, \textit{Cognitive Core}, and
\textit{Action Model}, forming a self-driven loop of
``execution--acquisition--evolution--execution'' that supports autonomous growth
without model retraining. Evolvable Memory constructs an external memory graph
through three-level hierarchical storage and a progressive
``record--summarize--abstract'' process, enabling experience to accumulate and
transfer across tasks. Cognitive Core unifies embodied cognition with
multimodal agent capabilities, grounding experience in the physical world and
consolidating reusable knowledge into skills. Focus-VLWA in the Action Model
turns world modeling into local dynamics supervision that directly supports
action learning. Together, the three modules allow execution to produce
experience and accumulated experience to improve subsequent actions, shifting
the embodied system from a ``train-and-freeze'' paradigm toward a
``deploy-and-evolve'' paradigm. Through hardware--software co-optimization with
M100, all three ME-Brain modules can also run independently on the edge chip,
providing an engineering foundation for scalable deployment in real physical
environments. This framework offers a feasible path toward embodied systems
capable of long-term autonomous growth in open physical environments.

\subsection{Future Work}
Future work will focus on the lifelong operation of the memory module and the evaluation of long-term embodied intelligence. For the memory system, we will further investigate memory compression, redundancy removal, and dynamic activity monitoring, using task relevance, access frequency, and temporal validity to identify and remove duplicated or low-value nodes and to consolidate information that no longer requires fine-grained representations. This will allow the system to preserve critical experiences and necessary associations while controlling storage and retrieval costs, preventing the continuous accumulation of historical data from degrading retrieval and maintenance efficiency and enabling memory to evolve throughout long-term interaction. Building on this capability, we will further explore secure on-device deployment of lifelong memory under resource-constrained settings, where user-related observations, interactions, and task experiences can be maintained and retrieved locally, together with improved data protection and memory management mechanisms. Through long-term interaction, the system can gradually accumulate knowledge of user preferences, behavioral patterns, and needs, and continuously update this knowledge as the user's routines and living context change, ultimately transforming long-term memory into a persistent and personalized service capability that enables embodied robots to better understand users and adapt to their evolving needs. In terms of evaluation, we will further conduct controlled comparisons and ablation studies covering long-term operation, environmental changes, and skill reuse, systematically examining the effects of memory and skill updates on task success rate, failure recovery capability, and computational and storage overhead.

%% file: sec/Contributiors.tex
\section{Contributions and Acknowledgments}
\label{sec:contributions}

\subsection*{Contributors}

\begin{itemize}
  \item \textbf{Memory:} Wei He, Hengtao Li, Chenfeng Wang, Zhongrui Yu
  \item \textbf{Cognition:} Xuhan Zhu, Maokui He, Zide Liu, Xiyue Zhang, Xianwei Mao, Chunpeng Zhou
  \item \textbf{Action:} Jia Shi, Yanze Xin, Jingwen Li, Jingxie Zheng, Sijie Zeng, Chenfeng Wang, Fan Lu, Zeyu Zhang, Shuai Guo, Hengxuan Zhang
  \item \textbf{System:} Zhongrui Yu, Wei He, Hengtao Li
  \item \textbf{Project Leader:} Pengfei Yu, Jia Shi, Yu Liu
  \item \textbf{Advisor:} Kun Zhan, Yan Xie
\end{itemize}

\subsection*{Acknowledgments}
Xueyang Zhang, Mofan Zhou, Bing Zhang, Yuying Chen, Shengyu Yao, Chang Ren, Chaoqun Du, Ming Li, Danlu Dong, Mingcui Wang, Jiao Deng, Yue Ma, Lingling Lu, Depeng Xu, Yimeng Li, Chen Liu, Liming Liu, Chunrui Wang

%% file: references.bib
@article{glm5,
  author    = {{GLM-5-Team} and Aohan Zeng and Xin Lv and others},
  title     = {{GLM-4.5}: Agentic, Reasoning, and Coding ({ARC}) Foundation Models},
  journal   = {arXiv preprint arXiv:2508.06471},
  year      = {2025},
}

@article{grpo,
  author    = {Zhihong Shao and Peiyi Wang and Qihao Zhu and Runxin Xu and Junxiao Song and Xiao Bi and Haowei Zhang and Mingchuan Zhang and Y. K. Li and Y. Wu and Daya Guo},
  title     = {{DeepSeekMath}: Pushing the Limits of Mathematical Reasoning in Open Language Models},
  journal   = {arXiv preprint arXiv:2402.03300},
  year      = {2024},
  url       = {https://arxiv.org/abs/2402.03300},
}

@article{liu2026simplemem,
  title={Simplemem: Efficient lifelong memory for llm agents},
  author={Liu, Jiaqi and Su, Yaofeng and Xia, Peng and Han, Siwei and Zheng, Zeyu and Xie, Cihang and Ding, Mingyu and Yao, Huaxiu},
  journal={arXiv preprint arXiv:2601.02553},
  year={2026}
}

@misc{chhikara2025mem0,
  author = {Prateek Chhikara and Dev Khant and Saket Aryan and Taranjeet Singh and Deshraj Yadav},
  title = {{Mem0: Building Production-Ready AI Agents with Scalable Long-Term Memory}},
  howpublished = {arXiv preprint arXiv:2504.19413},
  year = {2025},
  date = {2025-04-28},
  eprint = {2504.19413},
  archivePrefix = {arXiv},
  primaryClass = {cs.CL},
  doi = {10.48550/arXiv.2504.19413},
  url = {https://arxiv.org/abs/2504.19413},
  urldate = {2026-09-16},
}

@article{wang2025mirix,
  title={Mirix: Multi-agent memory system for llm-based agents},
  author={Wang, Yu and Chen, Xi},
  journal={arXiv preprint arXiv:2507.07957},
  year={2025}
}

@article{edge2024local,
  title={From local to global: A graph rag approach to query-focused summarization},
  author={Edge, Darren and Trinh, Ha and Cheng, Newman and Bradley, Joshua and Chao, Alex and Mody, Apurva and Truitt, Steven and Metropolitansky, Dasha and Ness, Robert Osazuwa and Larson, Jonathan},
  journal={arXiv preprint arXiv:2404.16130},
  year={2024}
}

@misc{li2025memos,
  author = {Zhiyu Li and Chenyang Xi and Chunyu Li and Ding Chen and Boyu Chen and Shichao Song and Simin Niu and Hanyu Wang and Jiawei Yang and Chen Tang and Qingchen Yu and Jihao Zhao and Yezhaohui Wang and Peng Liu and Zehao Lin and Pengyuan Wang and Jiahao Huo and Tianyi Chen and Kai Chen and Kehang Li and Zhen Tao and Huayi Lai and Hao Wu and Bo Tang and Zhengren Wang and Zhaoxin Fan and Ningyu Zhang and Linfeng Zhang and Junchi Yan and Mingchuan Yang and Tong Xu and Wei Xu and Huajun Chen and Haofen Wang and Hongkang Yang and Wentao Zhang and Zhi-Qin John Xu and Siheng Chen and Feiyu Xiong},
  title = {{MemOS: A Memory OS for AI System}},
  howpublished = {arXiv preprint arXiv:2507.03724},
  year = {2025},
  date = {2025-07-04},
  eprint = {2507.03724},
  archivePrefix = {arXiv},
  primaryClass = {cs.CL},
  doi = {10.48550/arXiv.2507.03724},
  url = {https://arxiv.org/abs/2507.03724},
  urldate = {2026-09-16},
}

@inproceedings{zhong2024memorybank,
  title={Memorybank: Enhancing large language models with long-term memory},
  author={Zhong, Wanjun and Guo, Lianghong and Gao, Qiqi and Ye, He and Wang, Yanlin},
  booktitle={Proceedings of the AAAI conference on artificial intelligence},
  volume={38},
  number={17},
  pages={19724--19731},
  year={2024}
}

@misc{shi2025memoryvla,
  author = {Hao Shi and Bin Xie and Yingfei Liu and Lin Sun and Fengrong Liu and Tiancai Wang and Erjin Zhou and Haoqiang Fan and Xiangyu Zhang and Gao Huang},
  title = {{MemoryVLA: Perceptual-Cognitive Memory in Vision-Language-Action Models for Robotic Manipulation}},
  howpublished = {arXiv preprint arXiv:2508.19236},
  year = {2025},
  date = {2025-08-26},
  eprint = {2508.19236},
  archivePrefix = {arXiv},
  primaryClass = {cs.RO},
  doi = {10.48550/arXiv.2508.19236},
  url = {https://arxiv.org/abs/2508.19236},
  urldate = {2026-09-16},
}

@misc{wang2026harness,
  author = {Qi Wang and Tianyi Wang and Chengyang Li and Shikun Ban and Yurun Chen and Yizhong Ge and Jason Qin and Chengtai Li and Wentao Zhu},
  title = {{Towards the Harness of Embodied Agents}},
  howpublished = {arXiv preprint arXiv:2608.11246},
  year = {2026},
  date = {2026-08-03},
  eprint = {2608.11246},
  archivePrefix = {arXiv},
  primaryClass = {cs.AI},
  doi = {10.48550/arXiv.2608.11246},
  url = {https://arxiv.org/abs/2608.11246},
  urldate = {2026-09-16},
}

@misc{torne2026mem,
  author = {Marcel Torne and Karl Pertsch and Homer Walke and Kyle Vedder and Suraj Nair and Brian Ichter and Allen Z. Ren and Haohuan Wang and Jiaming Tang and Kyle Stachowicz and Karan Dhabalia and Michael Equi and Quan Vuong and Jost Tobias Springenberg and Sergey Levine and Chelsea Finn and Danny Driess},
  title = {{MEM: Multi-Scale Embodied Memory for Vision Language Action Models}},
  howpublished = {arXiv preprint arXiv:2603.03596},
  year = {2026},
  date = {2026-03-04},
  eprint = {2603.03596},
  archivePrefix = {arXiv},
  primaryClass = {cs.RO},
  doi = {10.48550/arXiv.2603.03596},
  url = {https://arxiv.org/abs/2603.03596},
  urldate = {2026-09-16},
}

@misc{sridhar2025memer,
  author = {Ajay Sridhar and Jennifer Pan and Satvik Sharma and Chelsea Finn},
  title = {{MemER: Scaling Up Memory for Robot Control via Experience Retrieval}},
  howpublished = {arXiv preprint arXiv:2510.20328},
  year = {2025},
  date = {2025-10-23},
  eprint = {2510.20328},
  archivePrefix = {arXiv},
  primaryClass = {cs.RO},
  doi = {10.48550/arXiv.2510.20328},
  url = {https://arxiv.org/abs/2510.20328},
  urldate = {2026-09-16},
}

@misc{li2026roboclaw,
  author = {Ruiying Li and Yunlang Zhou and YuYao Zhu and Kylin Chen and Jingyuan Wang and Sukai Wang and Kongtao Hu and Minhui Yu and Bowen Jiang and Zhan Su and Jiayao Ma and Xin He and Yongjian Shen and Yang Yang and Guanghui Ren and Maoqing Yao and Wenhao Wang and Yao Mu},
  title = {{RoboClaw: An Agentic Framework for Scalable Long-Horizon Robotic Tasks}},
  howpublished = {arXiv preprint arXiv:2603.11558},
  year = {2026},
  date = {2026-03-12},
  eprint = {2603.11558},
  archivePrefix = {arXiv},
  primaryClass = {cs.RO},
  doi = {10.48550/arXiv.2603.11558},
  url = {https://arxiv.org/abs/2603.11558},
  urldate = {2026-09-16},
}

@misc{yang2026eventvla,
  author = {Ganlin Yang and Zhangzheng Tu and Yuqiang Yang and Sitong Mao and Junyi Dong and Tianxing Chen and Jiaqi Peng and Jing Xiong and Jiafei Cao and Jifeng Dai and Wengang Zhou and Yao Mu and Tai Wang},
  title = {{EventVLA: Event-Driven Visual Evidence Memory for Long-Horizon Vision-Language-Action Policies}},
  howpublished = {arXiv preprint arXiv:2606.20092},
  year = {2026},
  date = {2026-06-18},
  eprint = {2606.20092},
  archivePrefix = {arXiv},
  primaryClass = {cs.CV},
  doi = {10.48550/arXiv.2606.20092},
  url = {https://arxiv.org/abs/2606.20092},
  urldate = {2026-09-16},
}

@misc{liu2026embodiedscience,
  author = {Shaoshan Liu and Jie Tang},
  title = {{Embodied AI for Science}},
  howpublished = {Communications of the ACM, BLOG@CACM},
  year = {2026},
  date = {2026-02-04},
  url = {https://cacm.acm.org/blogcacm/embodied-ai-for-science/},
  urldate = {2026-09-16},
}

@misc{liu2026infrastructurefirst,
  author = {Shaoshan Liu and Jie Tang and Marwa S. Hassan and Mohamed H. Sharkawy and Moustafa M. G. Fouda and Tiewei Shang and Zixin Wang},
  title = {{Infrastructure First: Enabling Embodied AI for Science in the Global South}},
  howpublished = {arXiv preprint arXiv:2604.06722},
  year = {2026},
  date = {2026-04-08},
  eprint = {2604.06722},
  archivePrefix = {arXiv},
  primaryClass = {cs.CY},
  doi = {10.48550/arXiv.2604.06722},
  url = {https://arxiv.org/abs/2604.06722},
  urldate = {2026-09-16},
}

@article{yang2026vebrain15,
  author = {Ganlin Yang and Gen Luo and Ziyang Gong and Guanzhou Chen and Haonan Duan and Tianyi Zhang and Wengang Zhou and Yu Qiao and Wenhai Wang and Xizhou Zhu and Jifeng Dai},
  title = {{Visual Embodied Brain-1.5: Enhanced Perception, Spatial Reasoning and Robot Control in Spaces}},
  journal = {IEEE Transactions on Pattern Analysis and Machine Intelligence},
  year = {2026},
  date = {2026-08-27},
  pages = {1--18},
  doi = {10.1109/TPAMI.2026.3728098},
  url = {https://doi.org/10.1109/TPAMI.2026.3728098},
  urldate = {2026-09-16},
  note = {Early Access},
}

@misc{liu2026phyagentos,
  author = {Yang Liu and Weixing Chen and Xinshuai Song and Tao Pu and Siwen Mo and Yongjie Bai and Zihao Chen and Qianran Sun and Liruo Zhong and Ying Shen and Liang Lin},
  title = {{PhyAgentOS: A Self-Evolving Operating System for Embodied Agents with Decoupled Cognitive Planning and Physical Execution}},
  howpublished = {arXiv preprint arXiv:2607.16636},
  year = {2026},
  date = {2026-07-18},
  eprint = {2607.16636},
  archivePrefix = {arXiv},
  primaryClass = {cs.RO},
  doi = {10.48550/arXiv.2607.16636},
  url = {https://arxiv.org/abs/2607.16636},
  urldate = {2026-09-16},
}

@misc{acebrainteam2026acebrain05,
  author = {{ACE-Brain Team} and Ziyang Gong and Haoming Gu and Zehang Luo and Tianyi Zhang and Tao Tao and Yixiao Chi and Zhe Liu and Lingsi Zhu and Jingyuan Liu and Anke Tang and Songze Li and Yilun Kong and Ningjing Liu and Tianyu Zhu and Yunpeng Qing and Shuang Luo and Xiang Liu and Shi Fu and Dawei Nie and Sixiang Liu and Zhexi Wen and Feng Pan and Xiaofeng Wang and Zhi Hou and Chunxiao Liu and Xue Yang and Junchi Yan and Hengshuang Zhao and Dacheng Tao and Xiaogang Wang},
  title = {{ACE-Brain-0.5: A Unified Embodied Foundational Model for Physical Agentic AI}},
  howpublished = {arXiv preprint arXiv:2607.04426},
  year = {2026},
  date = {2026-07-05},
  eprint = {2607.04426},
  archivePrefix = {arXiv},
  primaryClass = {cs.RO},
  doi = {10.48550/arXiv.2607.04426},
  url = {https://arxiv.org/abs/2607.04426},
  urldate = {2026-09-16},
}

@misc{ju2026embodiskill,
  author = {Ruofei Ju and Xinrui Wang and Xin Ding and Yifan Yang and Hao Wu and Shiqi Jiang and Qianxi Zhang and Hao Wen and Xiangyu Li and Weijun Wang and Kun Li and Yunxin Liu and Haipeng Dai and Wei Wang and Ting Cao},
  title = {{EmbodiSkill: Skill-Aware Reflection for Self-Evolving Embodied Agents}},
  howpublished = {arXiv preprint arXiv:2605.10332},
  year = {2026},
  date = {2026-05-11},
  eprint = {2605.10332},
  archivePrefix = {arXiv},
  primaryClass = {cs.AI},
  doi = {10.48550/arXiv.2605.10332},
  url = {https://arxiv.org/abs/2605.10332},
  urldate = {2026-09-16},
}

@misc{wang2026phyai,
  author = {Chenghua Wang and Daliang Xu and Dongqi Cai and Duojin Sun and Hao Zhang and Haoze Qian and Huaiyuan Zhang and Jinshuo Cui and Junbo Cui and Kezhao Zhao and Longxi Gao and Mengwei Xu and Rongjie Yi and Ruixin Liu and Shangguang Wang and Tam Sikyuen and Tianyue Zhang and Weikai Xie and Xuanzhe Liu and Yingying Qin and Yiwen Lu and Yuan Yao and Yuezhi Zu and Yunhan Guo and Yuxin Zheng and Ziqi Guo},
  title = {{PhyAI: Real-Time Physical AI at the Edge, Scalable Rollouts in the Cloud}},
  howpublished = {arXiv preprint arXiv:2608.03682},
  year = {2026},
  date = {2026-08-04},
  eprint = {2608.03682},
  archivePrefix = {arXiv},
  primaryClass = {cs.AI},
  doi = {10.48550/arXiv.2608.03682},
  url = {https://arxiv.org/abs/2608.03682},
  urldate = {2026-09-16},
}

@misc{tan2025roboos,
  author = {Huajie Tan and Xiaoshuai Hao and Cheng Chi and Minglan Lin and Yaoxu Lyu and Mingyu Cao and Dong Liang and Zhuo Chen and Mengsi Lyu and Cheng Peng and Chenrui He and Yulong Ao and Yonghua Lin and Pengwei Wang and Zhongyuan Wang and Shanghang Zhang},
  title = {{RoboOS: A Hierarchical Embodied Framework for Cross-Embodiment and Multi-Agent Collaboration}},
  howpublished = {arXiv preprint arXiv:2505.03673},
  year = {2025},
  date = {2025-05-06},
  eprint = {2505.03673},
  archivePrefix = {arXiv},
  primaryClass = {cs.RO},
  doi = {10.48550/arXiv.2505.03673},
  url = {https://arxiv.org/abs/2505.03673},
  urldate = {2026-09-16},
}

@misc{kairosteam2026kairos,
  author = {{Kairos Team} and Fei Wang and Shan You and Qiming Zhang and Tao Huang and Zuoyi Fu and Zhisheng Zheng and Yunlong Xi and Feng Lv and Xiaoming Wu and Zeyu Liu and Cong Wan and Pu Li and Ruiqing Yang and Xiaoou Li and Wei Wang and Kangkang Zhu and Yuwei Zhang and Shi Fu and Zheng Zhang and Xiaoning Wu and Xuzeng Fan and Dacheng Tao and Xiaogang Wang},
  title = {{Kairos: A Regret-Aware Native World-Action Model Stack for Physical AI}},
  howpublished = {arXiv preprint arXiv:2606.16533},
  year = {2026},
  date = {2026-06-15},
  eprint = {2606.16533},
  archivePrefix = {arXiv},
  primaryClass = {cs.AI},
  doi = {10.48550/arXiv.2606.16533},
  url = {https://arxiv.org/abs/2606.16533},
  urldate = {2026-09-16},
}

@misc{chen2026host,
  author = {Guangyan Chen and Meiling Wang and Te Cui and Zichen Zhou and Qi Shao and Shalfun Li and Hang Su and Roy Gan and Hao Wang and Mengyin Fu and Yi Yang and Yufeng Yue},
  title = {{HOST: Robots Acquire Manipulation Skills in Seconds from a Single Human Video}},
  howpublished = {arXiv preprint arXiv:2607.20033},
  year = {2026},
  date = {2026-07-22},
  eprint = {2607.20033},
  archivePrefix = {arXiv},
  primaryClass = {cs.RO},
  doi = {10.48550/arXiv.2607.20033},
  url = {https://arxiv.org/abs/2607.20033},
  urldate = {2026-09-16},
}

@misc{liu2026guava,
  author = {Haowen Liu and Xirui Li and Shaoxiong Yao and Peng Shi and Tianyi Zhou and Jia-Bin Huang and Furong Huang and Jiayuan Mao},
  title = {{Guava: An Effective and Universal Harness for Embodied Manipulation}},
  howpublished = {arXiv preprint arXiv:2606.18363},
  year = {2026},
  date = {2026-06-16},
  eprint = {2606.18363},
  archivePrefix = {arXiv},
  primaryClass = {cs.RO},
  doi = {10.48550/arXiv.2606.18363},
  url = {https://arxiv.org/abs/2606.18363},
  urldate = {2026-09-16},
}

@misc{black2024pi0,
  author = {Kevin Black and Noah Brown and Danny Driess and Adnan Esmail and Michael Equi and Chelsea Finn and Niccolo Fusai and Lachy Groom and Karol Hausman and Brian Ichter and Szymon Jakubczak and Tim Jones and Liyiming Ke and Sergey Levine and Adrian Li-Bell and Mohith Mothukuri and Suraj Nair and Karl Pertsch and Lucy Xiaoyang Shi and James Tanner and Quan Vuong and Anna Walling and Haohuan Wang and Ury Zhilinsky},
  title = {{$\pi_0$: A Vision-Language-Action Flow Model for General Robot Control}},
  howpublished = {arXiv preprint arXiv:2410.24164v1},
  year = {2024},
  eprint = {2410.24164v1},
  archivePrefix = {arXiv},
  primaryClass = {cs.LG},
  url = {https://arxiv.org/abs/2410.24164v1},
  urldate = {2026-09-16},
}

@misc{physicalintelligence2025pi05,
  author = {{Physical Intelligence} and Kevin Black and Noah Brown and James Darpinian and Karan Dhabalia and Danny Driess and Adnan Esmail and Michael Equi and Chelsea Finn and Niccolo Fusai and Manuel Y. Galliker and Dibya Ghosh and Lachy Groom and Karol Hausman and Brian Ichter and Szymon Jakubczak and Tim Jones and Liyiming Ke and Devin LeBlanc and Sergey Levine and Adrian Li-Bell and Mohith Mothukuri and Suraj Nair and Karl Pertsch and Allen Z. Ren and Lucy Xiaoyang Shi and Laura Smith and Jost Tobias Springenberg and Kyle Stachowicz and James Tanner and Quan Vuong and Homer Walke and Anna Walling and Haohuan Wang and Lili Yu and Ury Zhilinsky},
  title = {{$\pi_{0.5}$: A Vision-Language-Action Model with Open-World Generalization}},
  howpublished = {arXiv preprint arXiv:2504.16054},
  year = {2025},
  eprint = {2504.16054},
  archivePrefix = {arXiv},
  primaryClass = {cs.LG},
  url = {https://arxiv.org/abs/2504.16054},
  urldate = {2026-09-16},
}

@misc{beyer2024paligemma,
  author = {Lucas Beyer and Andreas Steiner and Andr{\'e} Susano Pinto and Alexander Kolesnikov and Xiao Wang and Daniel Salz and Maxim Neumann and Ibrahim Alabdulmohsin and Michael Tschannen and Emanuele Bugliarello and Thomas Unterthiner and Daniel Keysers and Skanda Koppula and Fangyu Liu and Adam Grycner and Alexey Gritsenko and Neil Houlsby and Manoj Kumar and Keran Rong and Julian Eisenschlos and Rishabh Kabra and Matthias Bauer and Matko Bo{\v s}njak and Xi Chen and Matthias Minderer and Paul Voigtlaender and Ioana Bica and Ivana Balazevic and Joan Puigcerver and Pinelopi Papalampidi and Olivier Henaff and Xi Xiong and Radu Soricut and Jeremiah Harmsen and Xiaohua Zhai},
  title = {{PaliGemma: A Versatile 3B VLM for Transfer}},
  howpublished = {arXiv preprint arXiv:2407.07726},
  year = {2024},
  eprint = {2407.07726},
  archivePrefix = {arXiv},
  primaryClass = {cs.CV},
  url = {https://arxiv.org/abs/2407.07726},
  urldate = {2026-09-16},
}

@misc{ye2026dreamzero,
  author = {Seonghyeon Ye and Yunhao Ge and Kaiyuan Zheng and Shenyuan Gao and Sihyun Yu and George Kurian and Suneel Indupuru and You Liang Tan and Chuning Zhu and Jiannan Xiang and Ayaan Malik and Kyungmin Lee and William Liang and Nadun Ranawaka and Jiasheng Gu and Yinzhen Xu and Guanzhi Wang and Fengyuan Hu and Avnish Narayan and Johan Bjorck and Jing Wang and Gwanghyun Kim and Dantong Niu and Ruijie Zheng and Yuqi Xie and Jimmy Wu and Qi Wang and Ryan Julian and Danfei Xu and Yilun Du and Yevgen Chebotar and Scott Reed and Jan Kautz and Yuke Zhu and Linxi {``Jim''} Fan and Joel Jang},
  title = {{World Action Models Are Zero-shot Policies}},
  howpublished = {arXiv preprint arXiv:2602.15922},
  year = {2026},
  eprint = {2602.15922},
  archivePrefix = {arXiv},
  primaryClass = {cs.RO},
  url = {https://arxiv.org/abs/2602.15922},
  urldate = {2026-09-16},
}

@inproceedings{shi2023awe,
  author = {Lucy Xiaoyang Shi and Archit Sharma and Tony Z. Zhao and Chelsea Finn},
  title = {{Waypoint-Based Imitation Learning for Robotic Manipulation}},
  booktitle = {Proceedings of the 7th Conference on Robot Learning},
  series = {Proceedings of Machine Learning Research},
  volume = {229},
  pages = {2195--2209},
  publisher = {PMLR},
  year = {2023},
  url = {https://proceedings.mlr.press/v229/shi23b.html},
  urldate = {2026-09-16},
}

@misc{ye2026gigaworldpolicy,
  author = {Angen Ye and Boyuan Wang and Chaojun Ni and Guan Huang and Guosheng Zhao and Hao Li and Hengtao Li and Jie Li and Jindi Lv and Jingyu Liu and Min Cao and Peng Li and Qiuping Deng and Wenjun Mei and Xiaofeng Wang and Xinze Chen and Xinyu Zhou and Yang Wang and Yifan Chang and Yifan Li and Yukun Zhou and Yun Ye and Zhichao Liu and Zheng Zhu},
  title = {{GigaWorld-Policy: An Efficient Action-Centered World--Action Model}},
  howpublished = {arXiv preprint arXiv:2603.17240},
  year = {2026},
  eprint = {2603.17240},
  archivePrefix = {arXiv},
  primaryClass = {cs.CV},
  url = {https://arxiv.org/abs/2603.17240},
  urldate = {2026-09-16},
}

@misc{zhai2023siglip,
  author = {Xiaohua Zhai and Basil Mustafa and Alexander Kolesnikov and Lucas Beyer},
  title = {{Sigmoid Loss for Language Image Pre-Training}},
  howpublished = {arXiv preprint arXiv:2303.15343},
  year = {2023},
  eprint = {2303.15343},
  archivePrefix = {arXiv},
  primaryClass = {cs.CV},
  url = {https://arxiv.org/abs/2303.15343},
  urldate = {2026-09-16},
}

@inproceedings{perez2018film,
  author = {Ethan Perez and Florian Strub and Harm de Vries and Vincent Dumoulin and Aaron Courville},
  title = {{FiLM: Visual Reasoning with a General Conditioning Layer}},
  booktitle = {Proceedings of the AAAI Conference on Artificial Intelligence},
  volume = {32},
  number = {1},
  year = {2018},
  doi = {10.1609/aaai.v32i1.11671},
  url = {https://ojs.aaai.org/index.php/AAAI/article/view/11671},
  urldate = {2026-09-16},
}

@misc{zhang2019rmsnorm,
  author = {Biao Zhang and Rico Sennrich},
  title = {{Root Mean Square Layer Normalization}},
  howpublished = {arXiv preprint arXiv:1910.07467},
  year = {2019},
  eprint = {1910.07467},
  archivePrefix = {arXiv},
  primaryClass = {cs.LG},
  url = {https://arxiv.org/abs/1910.07467},
  urldate = {2026-09-16},
}

@inproceedings{lipman2023flowmatching,
  author = {Yaron Lipman and Ricky T. Q. Chen and Heli Ben-Hamu and Maximilian Nickel and Matt Le},
  title = {{Flow Matching for Generative Modeling}},
  booktitle = {International Conference on Learning Representations},
  year = {2023},
  url = {https://openreview.net/forum?id=PqvMRDCJT9t},
  urldate = {2026-09-16},
}

@misc{dai2026robomme,
  author = {Yinpei Dai and Hongze Fu and Jayjun Lee and Yuejiang Liu and Haoran Zhang and Jianing Yang and Chelsea Finn and Nima Fazeli and Joyce Chai},
  title = {{RoboMME: Benchmarking and Understanding Memory for Robotic Generalist Policies}},
  howpublished = {arXiv preprint arXiv:2603.04639},
  year = {2026},
  eprint = {2603.04639},
  archivePrefix = {arXiv},
  primaryClass = {cs.RO},
  url = {https://arxiv.org/abs/2603.04639},
  urldate = {2026-09-21},
}

@misc{chen2026robodojo,
  author = {Tianxing Chen and Yue Chen and Zixuan Li and Junyuan Tang and Kailun Su and Haoran Lu and Weijie Wan and Baijun Chen and Songling Liu and Haowen Yan and Honghao Su and Zhiyang Dou and Kaixuan Wang and Dandan Zhang and Yunze Liu and Yan Qin and Qiwei Liang and Qiwei Wu and Zijian Lin and Wenwei Lin and Yuran Wang and Minghua He and Tianshu Wu and Ruihai Wu and Jingquan Zhou and Kai-Chong Lei and Haibao Yu and Yuanfeng Ji and Weiyang Jin and Guanyu Lin and Xiaofan Li and Qi Xiong and Renjing Xu and Zhongyu Li and Wenhao Chai and Enze Xie and Ziwei Wang and Yao Mu and Hao Dong and Wojciech Matusik and Mingyu Ding and Wenbo Ding and Ping Luo and Masayoshi Tomizuka},
  title = {{RoboDojo: A Unified Sim-and-Real Benchmark for Comprehensive Evaluation of Generalist Robot Manipulation Policies}},
  howpublished = {arXiv preprint arXiv:2607.04434},
  year = {2026},
  eprint = {2607.04434},
  archivePrefix = {arXiv},
  primaryClass = {cs.RO},
  url = {https://arxiv.org/abs/2607.04434},
  urldate = {2026-09-21},
}

@article{yang2025agentic,
  title={Agentic robot: A brain-inspired framework for vision-language-action models in embodied agents},
  author={Yang, Zhejian and Chen, Yongchao and Zhou, Xueyang and Yan, Jiangyue and Song, Dingjie and Liu, Yinuo and Li, Yuting and Zhang, Yu and Zhou, Pan and Chen, Hechang and others},
  journal={arXiv preprint arXiv:2505.23450},
  year={2025}
}

@article{kim2024openvla,
  title={Openvla: An open-source vision-language-action model},
  author={Kim, Moo Jin and Pertsch, Karl and Karamcheti, Siddharth and Xiao, Ted and Balakrishna, Ashwin and Nair, Suraj and Rafailov, Rafael and Foster, Ethan and Lam, Grace and Sanketi, Pannag and others},
  journal={arXiv preprint arXiv:2406.09246},
  year={2024}
}

@article{kaelbling1998planning,
  title={Planning and acting in partially observable stochastic domains},
  author={Kaelbling, Leslie Pack and Littman, Michael L and Cassandra, Anthony R},
  journal={Artificial intelligence},
  volume={101},
  number={1-2},
  pages={99--134},
  year={1998},
  publisher={Elsevier}
}

@article{rynnbrain1_1,
  author    = {Kehan Li and Bohan Hou and Minghao Zhu and others},
  title     = {{RynnBrain 1.1}: Towards More Capable and Generalizable Embodied Foundation Model},
  journal   = {arXiv preprint arXiv:2607.17977},
  year      = {2026},
}

@article{dang2026rynnbrain,
  title={Rynnbrain: Open embodied foundation models},
  author={Dang, Ronghao and Guo, Jiayan and Hou, Bohan and Leng, Sicong and Li, Kehan and Li, Xin and Liu, Jiangpin and Mao, Yunxuan and Wang, Zhikai and Yuan, Yuqian and others},
  journal={arXiv preprint arXiv:2602.14979},
  year={2026}
}

@article{yuan2026embodied,
  title={Embodied-R1. 5: Evolving Physical Intelligence via Embodied Foundation Models},
  author={Yuan, Yifu and Huang, Yaoting and Yao, Xianze and Li, Yutong and Zhang, Shuoheng and Han, Linqi and Li, Pengyi and Sun, Jiangeng and Jia, Wenting and Zhang, Zhao and others},
  journal={arXiv preprint arXiv:2606.11324},
  year={2026}
}

@article{hy_embodied_vlm1_0,
  author    = {Ziyi Wang and Xumin Yu and Yongming Rao and others},
  title     = {{Hy-Embodied-VLM-1.0}: Efficient Physical-World Agents},
  journal   = {arXiv preprint arXiv:2607.12894},
  year      = {2026},
}

@article{bjorck2026vesta,
  title={Vesta: A Generalist Embodied Reasoning Model},
  author={Bjorck, Johan and Li, Zhiqi and Man, Yunze and Wang, Jing and Cheng, An-Chieh and Liu, Sifei and Wang, Shihao and Yu, Zhiding and Badki, Abhishek and Birchfield, Stan and others},
  journal={arXiv preprint arXiv:2606.20905},
  year={2026}
}

@misc{hao2025mimoembodiedxembodiedfoundationmodel,
      title={MiMo-Embodied: X-Embodied Foundation Model Technical Report}, 
      author={Xiaomi Embodied Intelligence Team},
      year={2025},
      eprint={2511.16518},
      archivePrefix={arXiv},
      primaryClass={cs.RO},
      url={https://arxiv.org/abs/2511.16518}, 
}

@misc{qwen38,
    title = {{Qwen3.8-Max}: A New Bar for Coding and Cowork},
    url = {https://qwen.ai/blog?id=qwen3.8},
    author = {{Qwen Team}},
    month = {August},
    year = {2026}
}

@misc{qwen2.5,
    title = {Qwen2.5: A Party of Foundation Models},
    url = {https://qwenlm.github.io/blog/qwen2.5/},
    author = {Qwen Team},
    month = {September},
    year = {2024}
}

@article{team2026mach,
  title={Mach-Mind-4-Flash Technical Report},
  author={Team, Foundation Model},
  journal={arXiv preprint arXiv:2607.09375},
  year={2026}
}

@article{team2026kimi,
  title={Kimi k3: Open frontier intelligence},
  author={Team, Kimi and Bai, Tongtong and Bai, Yifan and Bao, Yiping and Cai, Jianfeng and Cai, Xinyuan and Cao, Peizhou and Cao, Yuxuan and Chai, Ziwei and Charles, Y and others},
  journal={arXiv preprint arXiv:2607.24653},
  year={2026}
}

@misc{physbrain1.5,
      title={PhysBrain 1.5: From Vision-Language Models to Physical Foundation Models},
      author={DeepCybo Team and Yu Bin and Haipeng Cao and Zheng Chang and Kai Chen and Youning Chen and Kailin Deng and Yichao Du and Xiaotong Fu and Haoyang Ge and others},
      year={2026},
      eprint={2609.14973},
      archivePrefix={arXiv},
      primaryClass={cs.CV},
      url={https://arxiv.org/abs/2609.14973},
}
